\documentclass{article}
\usepackage{fancyhdr}

\usepackage{report}

\usepackage[utf8]{inputenc} 
\usepackage[T1]{fontenc}    
\usepackage{hyperref}       
\usepackage{url}            
\usepackage{booktabs}       
\usepackage{amsfonts}       
\usepackage{fancyhdr}       

\usepackage{nicefrac}       
\usepackage{microtype}      
\usepackage{graphicx}
\usepackage{pifont}
\usepackage{multirow}
\usepackage{CJKutf8}
\definecolor{darkmagenta}{rgb}{0.56, 0.0, 1.0}
\definecolor{softyellow}{rgb}{1.0, 0.92, 0.3} 
\definecolor{LightAquamarine}{rgb}{0.75, 1.0, 0.8} 
\definecolor{FireBrick}{RGB}{178,34,34}
\definecolor{MediumPurple}{RGB}{147,112,219}

\definecolor{uclablue}{rgb}{0.15, 0.45, 0.68}
\hypersetup{
    breaklinks,
    colorlinks=true,
    citecolor={darkmagenta},
    linkcolor={uclablue},
    urlcolor={uclablue}
}
\usepackage{wrapfig}
\usepackage{float}
\usepackage{subcaption}
\usepackage{placeins}
\usepackage{lipsum} 
\usepackage{tcolorbox}
\usepackage{amsmath}
\usepackage{amssymb}
\usepackage{fontawesome}
\usepackage{xspace}
\usepackage{enumitem}
\usepackage{multirow} 
\tcbuselibrary{breakable}
\usepackage{enumitem}
\usepackage{colortbl}
\usepackage{fancyhdr}

\usepackage{tcolorbox}
\usepackage{transparent}

\usepackage{microtype}
\usepackage{inconsolata}
\usepackage{graphicx}
\usepackage{amsmath}
\usepackage{amssymb}
\usepackage{booktabs}
\usepackage{listings}       
\usepackage{pifont}
\usepackage{multirow}
\usepackage{subcaption}
\usepackage{enumitem}
\usepackage{colortbl}
\usepackage{xcolor}
\usepackage{mdframed}

\usepackage[utf8]{inputenc}

\usepackage{microtype}

\usepackage{inconsolata}

\usepackage{tcolorbox}
\tcbuselibrary{skins, breakable}
\usepackage{tabularx}   
\usepackage{fontawesome} 
\usepackage{svg}         

\usepackage{xurl}
\hypersetup{colorlinks=true, urlcolor=blue, linkcolor=blue, citecolor=blue}
\usepackage{xcolor}

\newtcolorbox{promptbox}[1]{
    enhanced,
    breakable,    
    arc=5pt,                         
    boxrule=0.6pt,                   
    colframe=gray!60!black,          
    colbacktitle=gray!60!black,      
    colback=gray!12!white,           
    coltitle=white,                  
    fonttitle=\bfseries\small,       
    before skip=12pt,                
    after skip=6pt,                  
    toptitle=2pt,                    
    bottomtitle=2pt,                 
    left=12pt,                       
    right=12pt,                      
    top=6pt,                        
    bottom=6pt,                     
    title=#1
}

\definecolor{errorred}{RGB}{120, 30, 30}       
\definecolor{captionbg}{RGB}{248, 249, 250}   
\definecolor{captionborder}{RGB}{220, 225, 230}
\definecolor{analysisbg}{RGB}{255, 252, 240}  
\definecolor{analysisborder}{RGB}{230, 150, 50} 
\definecolor{trainedbg}{RGB}{246, 252, 246}     
\definecolor{trainedborder}{RGB}{85, 150, 105}  
\definecolor{gtbg}{RGB}{242, 246, 250}       
\definecolor{gtborder}{RGB}{70, 110, 150}    

\newtcolorbox{erroranalysiscontainer}[1]{
    enhanced,
    breakable,
    colback=white,
    colframe=errorred,
    arc=5pt,
    boxrule=1.2pt,
    title={\large\bfseries #1}, 
    colbacktitle=errorred,
    coltitle=white,
    before upper={\setlength{\parindent}{0pt}} 
}

\newtcolorbox{error_analysis_captionbox}[1]{
    enhanced,
    breakable,
    colback=captionbg,
    colframe=captionborder,
    arc=3pt,
    boxrule=0.5pt,
    title={\small\bfseries #1},
    colbacktitle=captionborder!80,
    coltitle=black,
    left=2pt, right=2pt, top=2pt, bottom=2pt,
    before skip=4pt, after skip=4pt
}

\newtcolorbox{error_analysis_analysisbox}[2]{
    enhanced,
    breakable,
    colback=#2,
    colframe=analysisborder,
    arc=3pt,
    boxrule=0.5pt,
    title={\small\bfseries #1},
    colbacktitle=analysisborder!60,
    coltitle=black,
    left=2pt, right=2pt, top=2pt, bottom=2pt,
    before skip=4pt, after skip=4pt
}

\newtcolorbox{error_analysis_trainedbox}[1]{
    enhanced,
    breakable,
    colback=trainedbg,                 
    colframe=trainedborder,     
    arc=3pt,
    boxrule=0.5pt,
    title={\small\bfseries #1}, 
    colbacktitle=trainedborder, 
    coltitle=white,             
    left=2pt, right=2pt, top=2pt, bottom=2pt,
    before skip=4pt, after skip=4pt
}

\newtcolorbox{error_analysis_gtbox}[1]{
    enhanced,
    breakable,
    colback=gtbg,                 
    colframe=gtborder,            
    arc=3pt,
    boxrule=0.5pt,
    title={\small\bfseries #1}, 
    colbacktitle=gtborder,        
    coltitle=white,               
    left=2pt, right=2pt, top=2pt, bottom=2pt,
    before skip=4pt, after skip=4pt
}

\lstdefinelanguage{json}{}

\definecolor{njuPurple}{RGB}{220,205,230}     
\definecolor{njuPurpleLight}{RGB}{250,245,252}   

\newtcolorbox{abstractbox}{
    colback=njuPurpleLight,   
    colframe=njuPurple,       
    boxrule=1pt,              
    arc=4mm,                  
    left=8pt,                 
    right=8pt,                
    top=8pt,                  
    bottom=8pt,               
    opacityback=0.95
}

\title{AVSCap: Orchestrating Audio-Visual Synergy for Omni-modal Video Captioning}

\author{
\textbf{Yanghai Wang$^{1*}$},
\textbf{Jiahao Wang$^{1*}$},
\textbf{Jiafu Tang$^{1*}$}, \\
\textbf{Yuanxing Zhang$^{2}$},
\textbf{Zhe Cao$^{1}$},
\textbf{Hanyan Bian$^{1}$}, 
\textbf{Zijie Zhang$^{1}$}, \\
\textbf{Weiliang Luo$^{1}$},
\textbf{Zhiyu Pan$^{1}$},
\textbf{Zixuan Dong$^{1}$},
\textbf{Jiaheng Liu$^{1}$},
\textbf{Zhaoxiang Zhang$^{3, \dagger}$} \\
\vspace{4mm}
{\normalsize $^1$ NJU-LINK Team, Nanjing University} \quad
{\normalsize $^2$ Kling Team, Kuaishou Technology} \\ 
{\normalsize $^3$  Institute of Automation, Chinese Academy of Sciences}  \\
\vspace{2mm}
\texttt{211300096@smail.nju.edu.cn}
\quad\quad\quad
\texttt{liujiaheng@nju.edu.cn} \\
}

\begin{document}

\maketitle
\let\oldthefootnote\thefootnote
\let\oldtheHfootnote\theHfootnote

\def\theHfootnote{authornote}
\let\thefootnote\relax\footnotetext{*~Equal Contribution. ~~$^\dagger$~Corresponding Author.}
\let\thefootnote\oldthefootnote
\let\theHfootnote\oldtheHfootnote

\begin{abstractbox}
\begin{center}
\textbf{\Large Abstract}
\end{center}
Omni-modal video captioning is not merely combining visual captioning with audio transcription: a useful caption must describe how visual actions, speech, music, and sound effects co-evolve. Existing large multimodal models often fail at this relational step, treating audio and visual streams as loosely coupled observations, relying on automatic speech recognition, and under-specifying non-speech sounds and their links to visual events. We present \textbf{AVSCap} \textsuperscript{\textit{\href{https://github.com/NJU-LINK/AVSCap}{a}}, \textit{\href{https://huggingface.co/datasets/NJU-LINK/AVSCapBench}{b}}}, a framework for audio-visual captioning centered on explicit cross-modal event binding. First, we construct \textbf{AVSCap-130K}, a tri-modal training corpus generated by a decoupled-then-fused pipeline that anchors visual and acoustic evidence before composing grounded omni-modal captions. Second, we train \textbf{AVSCap-7B}, a 7B captioner with a two-stage strategy: supervised fine-tuning establishes baseline capabilities, while sample-efficient reinforcement learning uses hybrid rewards to optimize acoustic completeness and audio-visual synergy. Our scaling analysis shows that reinforcement learning brings larger gains than increasing SFT data. Third, we introduce \textbf{AVSCapBench}, a benchmark that decomposes captions into visual, audio, and synergy events and evaluates them with fine-grained event recall. Experiments on AVSCapBench and external benchmarks show that AVSCap-7B improves non-speech audio coverage and cross-modal binding, delivering the best overall performance among evaluated open-source models.

\noindent\rule{0.32\linewidth}{0.4pt}

{\footnotesize
\textsuperscript{\textit{a}}\href{https://github.com/NJU-LINK/AVSCap}{https://github.com/NJU-LINK/AVSCap}\\
\textsuperscript{\textit{b}}\href{https://huggingface.co/datasets/NJU-LINK/AVSCapBench}{https://huggingface.co/datasets/NJU-LINK/AVSCapBench}
}

\end{abstractbox}

\begin{figure*}[t]
    \centering
    \includegraphics[width=\textwidth]{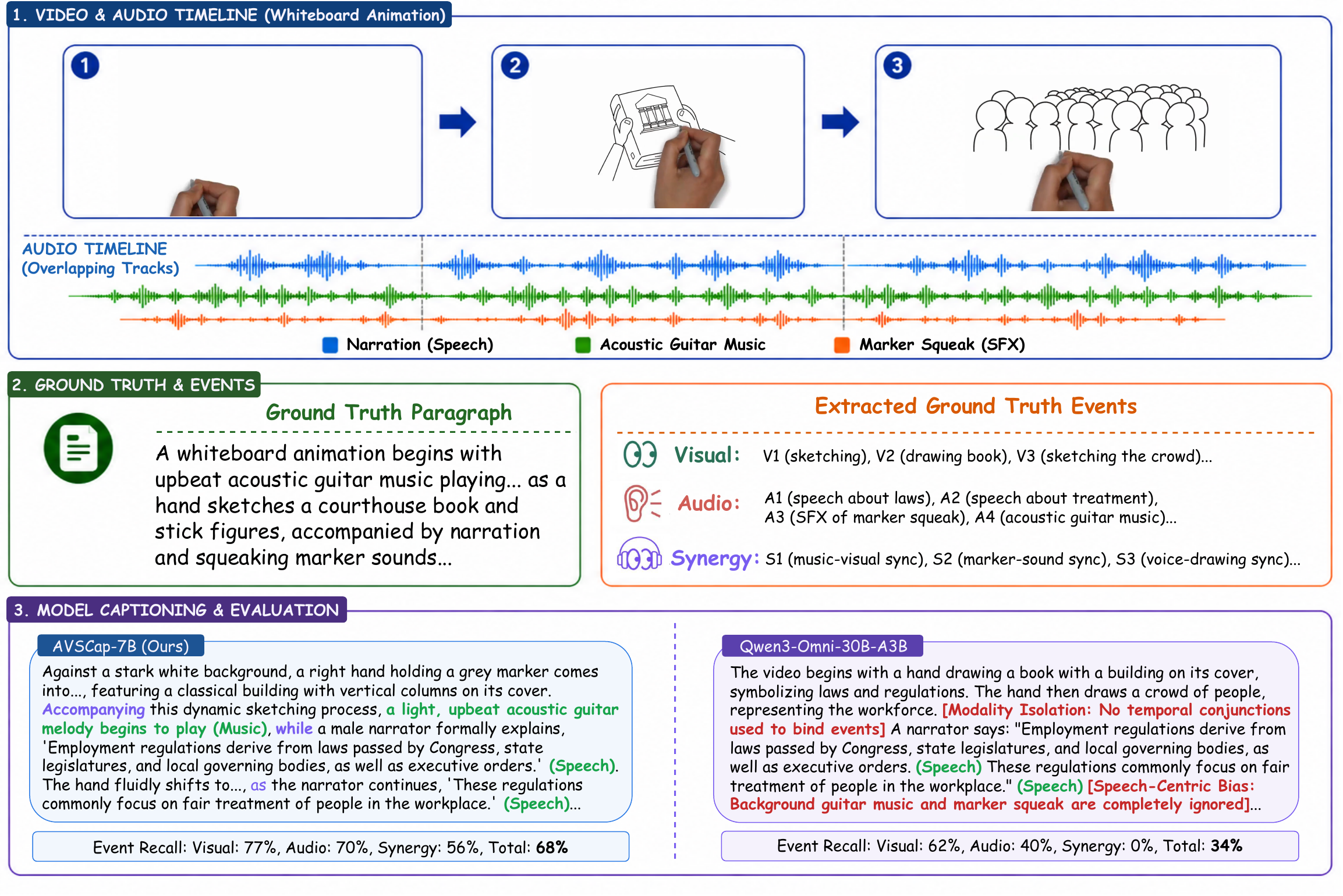} 
    \caption{Overview of our evaluation protocol and core bottlenecks in omni-modal video captioning. In the bottom captions, color-coded text highlights key aspects: green denotes audio components, purple represents synergy conjunctions, and red highlights major model limitations.}
    \label{fig:teaser}
\end{figure*}

\section{Introduction}
Large multimodal models (LMMs) have advanced video understanding toward omni-modal reasoning by jointly processing visual, audio, and textual signals~\citep{xu2025qwen25omnitechnicalreport, cheng2024videollama2, li2025baichuan, fu2025vita, liu2025ola}. Among related tasks, omni-modal video captioning remains fundamental: an ideal caption must recognize visual events, speech, sound effects, and music, and describe how they co-evolve over time~\citep{li2026asid,chen2026avocado}.

Despite recent progress, current models suffer from an illusion of integration (Figure~\ref{fig:teaser}), with two bottlenecks. First, \textbf{modality isolation}: models process audio and visual streams as weakly coupled channels. They may mention a visual action and a co-occurring sound but fail to express their temporal relations (e.g., ``while'', ``as''), yielding information-rich yet relation-poor captions. Second, \textbf{speech-centric bias}: acoustic representations are dominated by automatic speech recognition (ASR), leaving non-speech sounds (e.g., collisions, ambient effects, music) under-specified.

Current benchmarks also struggle to evaluate these issues~\citep{fu2025mme,li2024mvbench,wu2025ugc}. Existing protocols often reward independent recognition: models can score highly by outputting visual and audio descriptions without demonstrating cross-modal synergy. Acoustic evaluation also mainly targets ASR correctness, largely ignoring background sound effects and music.

To address these gaps, we present \textbf{AVSCap}, a framework for explicit cross-modal event binding. First, we construct \textbf{AVSCap-130K}, a tri-modal corpus generated via a decoupled-then-fused pipeline that anchors unimodal evidence before composing audio-visual captions. Second, we train \textbf{AVSCap-7B} (based on Qwen2.5-Omni) by SFT and Group Relative Policy Optimization (GRPO)~\citep{shao2024deepseekmath}. Guided by hybrid rewards, GRPO optimizes acoustic completeness and event binding. Finally, we introduce \textbf{AVSCapBench}, a human-curated benchmark decomposing captions into visual, audio, and synergistic events to evaluate audio sub-types and cross-modal synergy.

Our main contributions are:
\textbf{AVSCap-130K.} A tri-modal corpus of 130K orchestrated captions, providing explicit supervision for isolated perception and cross-modal grounding.
\textbf{AVSCapBench.} A human-annotated benchmark (1,226 videos) featuring a fine-grained, event-based matching protocol. It explicitly evaluates visual, audio, and synergistic events, preventing models from achieving high scores via modality isolation.
\textbf{AVSCap-7B.} A 7B captioner integrating SFT and GRPO to explicitly optimize acoustic completeness (including sound effects and music) and event binding.

\section{Related Work}
\label{sec:related_work}

\noindent\textbf{Audio-Visual Captioning.}
\label{subsec:rw_av_captioning}
The emergence of omni-modal models~\citep{comanici2025gemini25, xu2025qwen25omnitechnicalreport, ai2025ming} has shifted video understanding from vision-centric perception to joint audio-visual modeling~\citep{chen2026avocado}. Representative works include UGC-VideoCaptioner~\citep{wu2025ugc} and video-SALMONN-2~\citep{tang2025salmonn2} for multimodal integration, AVoCaDO~\citep{chen2026avocado} for audiovisual temporal coherence, Omni-Captioner~\citep{ma2026omnicaptioner} and ASID-Captioner~\citep{li2026asid} for detailed perception, TimeChat-Captioner~\citep{yao2026timechat} for structured multi-scene scripting, and OmniScript~\citep{pu2026omniscript} for hierarchical script generation. While these methods improve audio-visual captioning, they do not center data construction, training, and evaluation around explicit event-level audio-visual binding. In contrast, AVSCap optimizes cross-modal synergy through a decoupled-then-fused data engine that anchors unimodal evidence before composing audio-visual captions.

\noindent\textbf{RL for Video Captioning.}
\label{subsec:rw_rl_captioning}
RL~\citep{schulman2017proximal, guo2025deepseek, zheng2025group, gao2025soft} has become an important paradigm for aligning multimodal models with explicit objectives. CapRL~\citep{xing2025caprl} introduces verifiable rewards for caption generation, and VideoCap-R1~\citep{meng2025videocap} uses GRPO to elicit structured thinking before captioning. AVoCaDO extends GRPO with content coverage and length regularization rewards, while TimeChat-Captioner trains its policy with task-specific rewards to align dense captions with fine-grained temporal boundaries. We adopt a regex-guided GRPO strategy with hybrid rewards to jointly optimize acoustic completeness and cross-modal event binding.

\begin{figure*}[t]
    \centering
    \includegraphics[width=\textwidth]{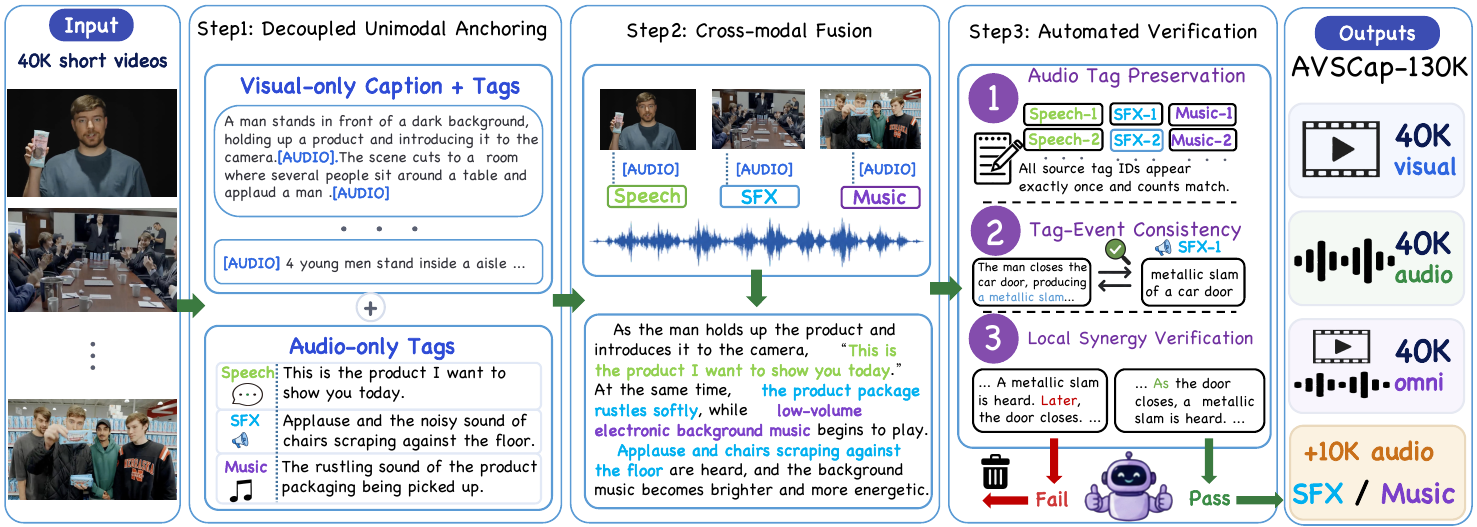} 
    \caption{Overview of the AVSCap-130K data construction pipeline.}
    \label{fig:trainset}
\end{figure*}

\section{AVSCap}
\label{sec:methodology}

\subsection{Data Construction Pipeline}
\label{sec:training_data}

\subsubsection{Task Definition}
\label{subsec:task_definition}

Existing video captioning tasks mainly emphasize isolated visual or audio accuracy. We instead define a high-quality omni-modal caption by three criteria: (1) \textbf{Acoustic Completeness}, covering speech \texttt{(Speech)}, sound effects \texttt{(SFX)}, and music; (2) \textbf{Visual Completeness}, describing environments, characters, actions, object interactions, camera motion, and OCR; and (3) \textbf{Audio-Visual Synergy}, binding audio and visual events through coherent cross-modal relations, e.g., linking ``a hammer falls'' with ``a striking sound \texttt{(SFX)}''.

\subsubsection{Orchestrated Data Engine}
\label{sec:data_engine}
To train \textbf{AVSCap-7B}, we construct \textbf{AVSCap-130K}, a 40K-video corpus with temporally grounded omni-modal captions. Videos are collected from AVoCaDO-107K~\citep{chen2026avocado}, ASID-1M~\citep{li2026asid}, FineVideo~\citep{Farre2024FineVideo}, TimeChatCap-40K~\citep{yao2026timechat}, and Movie101~\citep{yue2023movie101}, and filtered to clips shorter than 2.5 minutes. As shown in Figure~\ref{fig:trainset}, our decoupled-then-fused data engine has three stages: unimodal anchoring, cross-modal orchestration, and automated verification. Prompts used in this section are provided in Appendix~\ref{subsec:prompts_training_set}.

\paragraph{Step 1: Decoupled Unimodal Anchoring}
\label{step1}

To reduce cross-modal interference during initial perception, we adopt a decoupled unimodal anchoring strategy using Gemini-3-Flash~\citep{Google2026Gemini30Flash}. For the visual stream, the parser extracts scene-level attributes, including environments, characters, actions, object interactions, camera motion, and OCR text. When an audio cue is detected, the parser inserts an empty placeholder to indicate that the current visual segment has corresponding audio information, without describing the acoustic content itself. This preserves the visual-only nature of the caption while retaining temporal anchors for later cross-modal fusion.

For the audio stream, the parser separately extracts three acoustic categories: human speech \texttt{(Speech)}, sound effects \texttt{(SFX)}, and background music \texttt{(Music)}. To standardize acoustic descriptions, SFX captions follow the annotation style of AudioCaps~\citep{kim-NAACL-HLT-2019}, while music captions follow the descriptive format of MusicCaps~\citep{agostinelli2023musiclm}.

\paragraph{Step 2: Cross-modal Fusion }

Given the unimodal anchors, we use Gemini-3-Flash to synthesize a coherent omni-modal caption. The fusion module aligns each visual placeholder with its corresponding audio description and inserts the audio content at the matched temporal position. To make cross-modal relations explicit, the fused caption uses temporal conjunctions such as ``As'', ``While'', ``Simultaneously'', and ``Accompanied by''. The original audio tags \texttt{(Speech)}, \texttt{(SFX)}, and \texttt{(Music)} are retained for later verification.

\paragraph{Step 3: Automated Verification}

To reduce hallucinations, duplication, and information loss during fusion, we apply a three-stage verification pipeline combining deterministic tag checks with local semantic validation.

\noindent \textbf{Audio Tag Preservation.}
Since audio captions are chronologically ordered, we assign each audio event a unique tag ID before fusion, such as \texttt{(Speech-1)}, \texttt{(SFX-1)}, and \texttt{(Music-1)}. After fusion, we verify that every source tag ID appears exactly once and that the retained tag count matches the number of source audio events for each type. Samples with missing, duplicated, or mismatched tags are discarded, as they indicate omitted, repeated, or misplaced audio events.

\noindent \textbf{Tag-Event Consistency Check.}
Each retained audio tag must be attached to a local caption sentence and correspond to an integrated audio event. We use Qwen3-32B~\citep{yang2025qwen3} to verify whether the fused audio event remains semantically consistent with its source audio caption. Samples with unmatched tags or altered audio meanings are discarded to filter hallucinated, duplicated, or temporally misplaced audio insertions.

\noindent \textbf{Local Synergy Verification.}
We further use Qwen3-32B to check whether each tag-anchored sentence contains both the audio event and its corresponding visual context, expressed through temporal or associative cues such as ``while'', ``as'', or ``accompanied by''. Captions failing this check are removed, since they may preserve both modalities without forming valid cross-modal event binding.

\subsection{Training AVSCap-7B}
\label{sec:model_training}

\textbf{AVSCap-7B} is built on Qwen2.5-Omni-7B and trained in two stages: SFT establishes baseline captioning, and GRPO improves acoustic completeness and audio-visual event binding. Additional details are provided in Appendix~\ref{sec:training_details}.
\subsubsection{Stage 1: Supervised Fine-Tuning (SFT)}
We fine-tune Qwen2.5-Omni-7B with a multi-grained SFT dataset. Each video in \textbf{AVSCap-130K} provides three annotations: a visual caption, an audio caption, and a synergistic omni-modal caption. These prompt-response pairs enable both unimodal perception and cross-modal orchestration.

To improve non-speech audio understanding, we construct 10,000 audio-centric captions from AudioCaps~\citep{kim-NAACL-HLT-2019} and MusicCaps~\citep{agostinelli2023musiclm}. Their raw audio tracks are processed with the same audio-stream pipeline in Section~\ref{step1} to generate detailed \texttt{(SFX)} and \texttt{(Music)} captions. All video and audio captions are then jointly shuffled for unified SFT training. Ablations are provided in Appendix~\ref{subsec:audio_augmentation}.

\subsubsection{Stage 2: GRPO Optimization}

Although SFT provides strong captioning ability, the model still produces occasional repetitive outputs and weak audio-visual event binding. We further optimize \textbf{AVSCap-7B} with Group Relative Policy Optimization (GRPO) on 2,000 additional videos that do not overlap with the SFT training set. Their reference captions are constructed with the same data pipeline as \textbf{AVSCap-130K}. For each training instance, the policy samples a group of candidate captions, and optimization is guided by three rewards for length control, speech preservation, and audio-visual consistency.

\noindent \textbf{Reward 1: Length Regularization ($R_{\mathrm{len}}$).}
To prevent short outputs and repetition collapse, we assign full reward to captions within a valid length range and zero otherwise:
\begin{equation}
R_{\mathrm{len}} =
\mathbb{I}\left(\tau_{\min} \leq L_{\mathrm{gen}} \leq \tau_{\max}\right),
\end{equation}
where $L_{\mathrm{gen}}$ is the generated caption length. We set $\tau_{\min}=200$ and $\tau_{\max}=2048$ based on caption-length statistics and context-budget analysis, with details provided in Appendix~\ref{app:length_analysis}.

\noindent \textbf{Reward 2: Regex-Anchored Speech Recall ($R_{\mathrm{sp}}$).}
To encourage faithful preservation of speech content without expensive judge calls, we extract all \texttt{(Speech)} segments via regex matching and compare them with the reference speech segments. Before matching, punctuation is removed from both sides. For each aligned speech pair, we compute an LCS-based recall score:
\begin{equation}
R_{\mathrm{sp}} =
\frac{1}{M}\sum_{i=1}^{M}
\frac{
\mathrm{LCS}\left(s_i^{\mathrm{gen}}, s_i^{\mathrm{ref}}\right)
}{
\left|s_i^{\mathrm{ref}}\right|
},
\end{equation}
where $M$ is the number of reference speech segments, $s_i^{\mathrm{gen}}$ and $s_i^{\mathrm{ref}}$ denote the generated and reference speech segments after tag-based alignment, and $\mathrm{LCS}( , )$ denotes the length of their longest common subsequence. This reward penalizes omitted, reordered, or substantially rewritten speech while remaining fully rule-based.

\noindent \textbf{Reward 3: Cross-modal Synergy Recall ($R_{\mathrm{syn}}$).}
To optimize the correctness and consistency of audio-visual event binding, we decompose the reference caption into synergy events and use GPT-5~\citep{singh2025gpt5} to judge whether each event is covered by the generated caption. The reward is computed as event-level recall:
\begin{equation}
R_{\mathrm{syn}} =
\frac{
\left|\mathcal{E}_{\mathrm{covered}}\right|
}{
\left|\mathcal{E}_{\mathrm{syn}}\right|
},
\end{equation}
where $\mathcal{E}_{\mathrm{syn}}$ is the set of reference synergy events and $\mathcal{E}_{\mathrm{covered}}$ is the subset covered by the generated caption. This reward encourages preserving both modalities and their temporal correspondence.

\noindent \textbf{Total Reward.}
The final reward is the sum of the three components:
\begin{equation}
R_{\mathrm{total}} = R_{\mathrm{len}} + R_{\mathrm{sp}} + R_{\mathrm{syn}}.
\end{equation}

\begin{table}[t]
\centering
\begin{tabular}{lcccccc}
\toprule
\textbf{Benchmark} & \textbf{\#Videos} & \textbf{Avg. D} & \textbf{A} & \textbf{V} & \textbf{AVS} & \textbf{Sub-T} \\
\midrule
UGC-VidCap & 1,000 & 23.9s & \ding{51} & \ding{51} & \ding{55} & \ding{55} \\
Omni-Cloze & 2,340 & 34.2s & \ding{51} & \ding{51} & \ding{51} & \ding{55} \\
v-SALMONN2 & 483 & 50.8s & \ding{51} & \ding{51} & \ding{55} & \ding{55} \\
\midrule
\textbf{AVSCapBench (Ours)} & \textbf{1,226} & \textbf{60.1s} & \ding{51} & \ding{51} & \ding{51} & \ding{51} \\
\bottomrule
\end{tabular}
\caption{Comparison of audio-visual video captioning benchmarks. Avg. D: Average Duration; A: Audio; V: Visual; AVS: Audio-Visual Synergy; Sub-T: Audio Sub-types. ``v-SALMONN2'' is abbreviated for the video-SALMONN-2-testset. \ding{51}: supported; \ding{55}: not supported.}
\label{tab:benchmark_comparison}
\end{table}

\begin{figure*}[t]
  \centering
  \includegraphics[width=\textwidth]{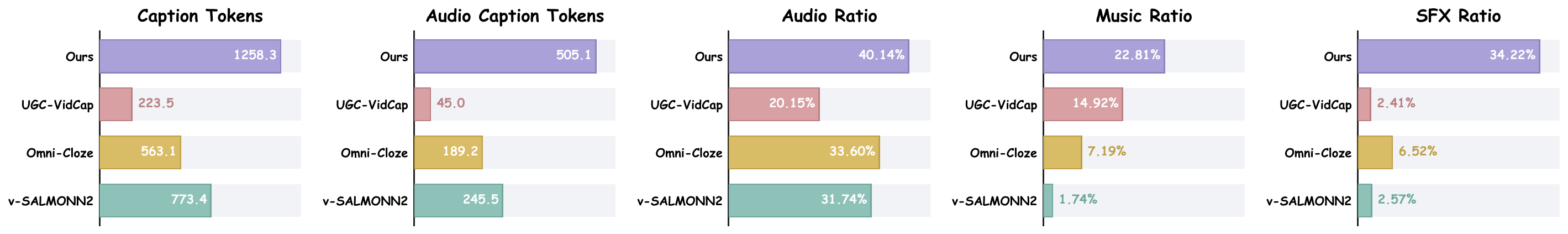} 
  \caption{Comparison of caption characteristics across benchmarks. Metrics include average token lengths for total and audio-specific descriptions (Caption/Audio Tokens), the ratio of audio tokens (Audio Ratio), and internal distribution of music and sound effects (Music/SFX Ratio). v-SALMONN2 refers to video-SALMONN-2 testset.}
  \label{fig:benchmark_comparison}
\end{figure*}

\subsection{The AVSCapBench}
\label{sec:benchmark}

To rigorously evaluate the capabilities of omni-modal models in video understanding and audio-visual synergy, we introduce the AVSCapBench.

\subsubsection{Benchmark Construction}

\begin{figure*}[t]
    \centering
    \includegraphics[width=\textwidth]{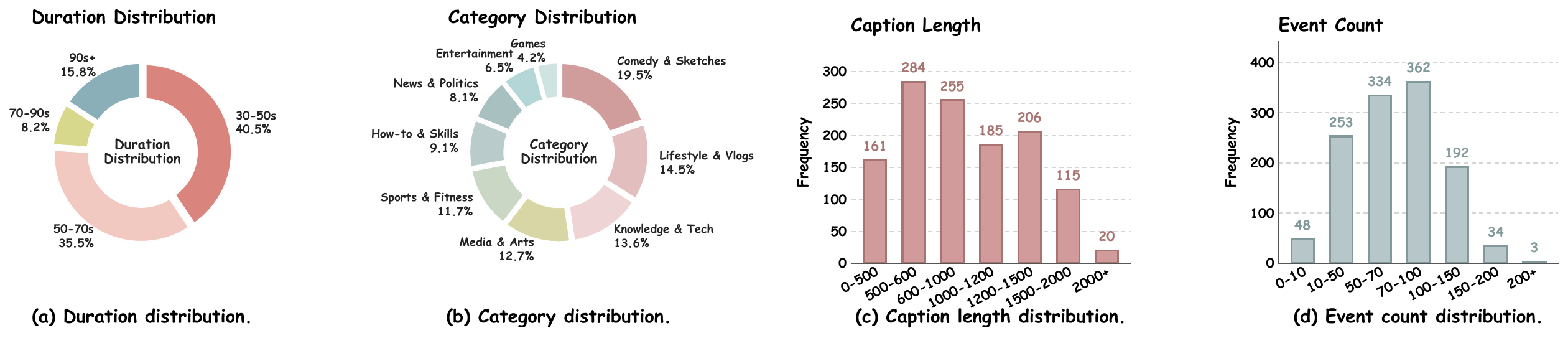} 
    \caption{Statistics of AVSCapBench. (a) Videos span durations from 30 to 120 seconds. (b) Diverse categories. (c) Caption lengths vary across the dataset. (d) Distribution of atomic events per video.}
    \label{fig:benchmark_stats}
\end{figure*}

\textbf{AVSCapBench} consists of 1,226 manually annotated video clips collected from YouTube, TikTok, and Video-MME~\citep{fu2025mme}, lasting 30 to 120 seconds. It covers diverse domains, including movies, vlogs, gaming, sports, and news. A complete annotation case is provided in Appendix~\ref{sec:appendix_example}. Table~\ref{tab:benchmark_comparison} and Figure~\ref{fig:benchmark_comparison} compare AVSCapBench with existing benchmarks, and Figure~\ref{fig:benchmark_stats} summarizes its statistics.
We construct AVSCapBench through a three-stage human-in-the-loop pipeline.

\noindent \textbf{Automated Segmentation.}
We first use Gemini-3-Flash~\citep{Google2026Gemini30Flash} to identify segmentation timestamps that preserve both visual coherence and audio continuity. The original videos are then divided into shorter clips for annotation.

\noindent \textbf{Segment-level Human Annotation.}
Human annotators independently caption each clip, covering visual content like scenes, characters, actions, objects, and OCR text alongside speech, sound effects, and music. Annotators are instructed to place audio descriptions near their corresponding visual events and to ensure that each caption reflects audio-visual consistency throughout the segment.

\noindent \textbf{Merging and Cross-Verification.}
Segment-level captions are sequentially merged into dense full-video captions. A second group of annotators then cross-checks them to identify omissions, hallucinations, and incorrect audio-visual alignments. To stress-test fine-grained acoustic coverage, we retain only samples that contain all three audio categories---\texttt{Speech}, \texttt{SFX}, and \texttt{Music}; samples that do not satisfy this requirement are discarded.

\subsubsection{Evaluation Protocol}
\label{subsec:evaluation_protocol}

Inspired by DREAM-1K~\citep{wang2024tarsier}, we adopt a fine-grained event-based matching protocol, as illustrated in Figure~\ref{fig:teaser}. We utilize GPT-5~\citep{singh2025gpt5} to decompose each ground-truth caption directly into a structured set of atomic events $\mathcal{E}$ categorized across three distinct modality types. \textbf{Visual events} $\mathcal{E}_{visual}$ describe objective visual content, including entities, actions, interactions, and scene states. \textbf{Audio events} $\mathcal{E}_{audio}$ cover auditory information, including speech, sound effects, and background music, organized in chronological order. \textbf{Synergy events} $\mathcal{E}_{synergy}$ bind a visual event with its corresponding audio cue when the two are temporally aligned, thereby capturing cross-modal audio-visual relations. We empirically demonstrate the limitations of traditional n-gram overlap metrics and justify our transition to event-based recall in Appendix~\ref{sec:appendix_traditional_metrics}.

During evaluation, we use Gemini-3.1-Pro~\citep{Google2026Gemini31Pro} as the judge model. For each event $e_i \in \mathcal{E}$, the judge determines whether the model-generated caption $\hat{C}$ semantically covers the event. We then compute recall independently for each event type:
\begin{equation}
\text{Recall}_{\text{type}} =
\frac{1}{|\mathcal{E}_{\text{type}}|}
\sum_{e_i \in \mathcal{E}_{\text{type}}}
h(e_i, \hat{C}),
\end{equation}
where $h(e_i,\hat{C})=1$ if the judge determines that $\hat{C}$ covers event $e_i$, and $h(e_i,\hat{C})=0$ otherwise. 
The event type is defined as $\text{type} \in \{\text{visual}, \text{audio}, \text{synergy}\}$. Since the numbers of events differ across modality types, the overall benchmark score is computed as the event-count-weighted average of the three recall metrics. The complete set of system instructions and evaluation prompts is detailed in Appendix~\ref{sec:appendix_prompts}.

\section{Experiments}
\label{sec:experiments}
\subsection{Main Results}
We evaluate 13 leading omni-modal models, including Gemini-3-Pro~\citep{Google2026Gemini30Pro}, Gemini-3-Flash~\citep{Google2026Gemini30Flash}, Qwen3-Omni~\citep{xu2025qwen3omnitechnicalreport}, Qwen2.5-Omni~\citep{xu2025qwen25omnitechnicalreport}, ARC-Hunyuan-Video~\citep{ge2025arc}, HumanOmniV2~\citep{yang2025humanomniv2}, MiniCPM-o~\citep{yao2024minicpm}, video-SALMONN-2~\citep{tang2025salmonn2}, ASID-Captioner~\citep{li2026asid}, AVoCaDO~\citep{chen2026avocado}, and UGC-VideoCaptioner~\citep{wu2025ugc}. We provide the detailed evaluation settings and an extended set of evaluation results in Appendix~\ref{sec:evaluation_settings} and~\ref{sec:complete_main_results}, respectively. The main results in Table~\ref{tab:main_results} yield several key observations: (1)~\textbf{Audio-visual synergy remains the weakest capability.} Synergy Recall is consistently the lowest dimension across all models. Open-source models often exhibit near-zero synergy despite strong unimodal perception, indicating a severe lack of temporal alignment. (2)~\textbf{Specialized models outperform general omni-modal models.} Task-specific captioners (e.g., AVoCaDO) surpass general omni-modal models (e.g., Qwen2.5-Omni), underscoring the value of task-oriented training. (3)~\textbf{Audio understanding is highly inconsistent} across open-source systems, with music and SFX remaining particularly challenging. (4)~\textbf{AVSCap-7B achieves competitive performance}, outperforming open-source baselines and approaching commercial systems, validating the effectiveness of our curated data and GRPO-based optimization.


\begin{table*}[t]
\centering
\resizebox{0.85\textwidth}{!}{
\begin{tabular}{lcccccccl}
\toprule
\multirow{2}{*}{\textbf{Model}}
& \multirow{2}{*}{\textbf{Visual}}
& \multicolumn{4}{c}{\hspace{0.5em}\textbf{Audio}\hspace{0.5em}}
& \multirow{2}{*}{\textbf{Synergy}}
& \multirow{2}{*}{\textbf{Total}} \\
\cmidrule(lr){3-6}
&
& \textbf{Speech}
& \textbf{Music}
& \textbf{SFX}
& \textbf{Overall}
& & \\
\midrule
\multicolumn{8}{c}{\textit{Closed-Source Models}} \\
\midrule
Gemini-3-Pro & 60.43 & 79.81 & 39.52 & 27.77 & 71.29 & 48.88 & 60.97 \\
Gemini-3-Flash & 58.14 & 79.78 & 39.46 & 32.34 & 72.65 & 48.94 & 60.54 \\
\midrule
\multicolumn{8}{c}{\textit{Open-Source Models}} \\
\midrule
AVoCaDO-7B & 50.59 & 70.42 & 38.71 & 19.25 & 61.07 & 29.13 & 49.31 \\
ASID-Captioner-7B & 47.42 & 68.73 & 30.50 & 17.91 & 59.02 & 24.84 & 45.94 \\
ASID-Captioner-3B & 43.63 & 66.95 & 27.06 & 17.31 & 57.53 & 21.36 & 43.03 \\
Qwen3-Omni-30B-A3B-Instruct & 41.85 & 49.08 & 9.34 & 8.68 & 39.17 & 16.19 & 35.29 \\
video-SALMONN-2-7B & 39.05 & 46.76 & 13.76 & 8.71 & 36.52 & 12.43 & 32.02 \\
UGC-VideoCaptioner-3B & 33.24 & 21.30 & 22.00 & 11.48 & 20.77 & 10.43 & 24.24 \\
Qwen2.5-Omni-7B & 34.78 & 13.92 & 4.02 & 7.22 & 13.71 & 7.00 & 21.53 \\
ARC-Hunyuan-Video-7B & 20.68 & 16.49 & 3.93 & 1.97 & 11.41 & 4.52 & 14.49 \\
HumanOmniV2-7B & 27.78 & 4.60 & 1.58 & 2.46 & 4.41 & 2.42 & 14.10 \\
MiniCPM-o-2.6-8B & 24.61 & 6.75 & 3.31 & 3.92 & 6.13 & 3.78 & 13.66 \\
\midrule
\textbf{AVSCap-7B (Ours)} & \textbf{59.33} & \textbf{69.45} & \textbf{40.36} & \textbf{30.82} & \textbf{64.30} & \textbf{57.70} & \textbf{60.44} \\
\bottomrule
\end{tabular}
}
\caption{Main results on AVSCapBench. All values are Recall (\%).}
\label{tab:main_results}
\end{table*}

\subsection{Cross-Benchmark Evaluation}
\label{subsec:cross_benchmark}

To assess generalization, we evaluate AVSCap-7B on UGC-VideoCap~\citep{wu2025ugc}, Daily-Omni~\citep{zhou2025dailyomni}, and Omni-Cloze~\citep{ma2026omnicaptioner}. Detailed descriptions of these external benchmarks can be found in Appendix~\ref{sec:details_of_benchmarks}. As shown in Table~\ref{tab:cross_benchmark_main}, AVSCap-7B demonstrates robust transferability, outperforming all open-source baselines and approaching proprietary models on both UGC-VideoCap and Daily-Omni. Furthermore, Table~\ref{tab:cross_benchmark_cloze} confirms its superiority on Omni-Cloze, particularly in the Audio-Visual subset, validating that our synergy-focused training successfully transfers to dense cross-modal reasoning tasks.

\begin{table}[t]
\centering
\begin{tabular}{l cccc c}
\toprule
& \multicolumn{4}{c}{\textbf{UGC-VideoCap}} & \textbf{Caption-to-QA} \\
\cmidrule(lr){2-5} \cmidrule(lr){6-6}
\textbf{Model} & Aud. $\uparrow$ & Vis. $\uparrow$ & Det. $\uparrow$ & Avg. $\uparrow$ & Daily-Omni $\uparrow$ \\
\midrule
\multicolumn{6}{c}{\textit{Closed-Source Commercial Models}} \\
\midrule
Gemini-3-Pro & 80.4 & 84.7 & 80.6 & 81.9 & 66.4 \\
Gemini-2.5-Pro & 69.5 & 74.7 & 73.7 & 72.6 & 60.2 \\
\midrule
\multicolumn{6}{c}{\textit{Open-Source Models}} \\
\midrule
HumanOmniV2-7B & 45.6 & 66.3 & 59.5 & 57.1 & 8.2 \\
ARC-Hunyuan-Video-7B & 52.7 & 56.0 & 55.8 & 54.8 & 8.6 \\
MiniCPM-o-2.6-8B & 38.6 & 68.5 & 57.7 & 54.9 & 9.8 \\
Qwen2.5-Omni-7B & 46.9 & 66.1 & 60.0 & 57.7 & 13.4 \\
UGC-VideoCaptioner-3B & 61.4 & 58.4 & 57.5 & 59.1 & 17.0 \\
video-SALMONN-2-7B & 61.8 & 71.4 & 68.5 & 67.2 & 29.9 \\
Qwen3-Omni-30B-A3B & 67.5 & 74.8 & 72.3 & 71.5 & 17.5 \\
AVoCaDO-7B & 73.0 & 74.6 & 71.8 & 73.2 & 50.1 \\
ASID-Captioner-3B & 78.6 & 84.8 & 80.2 & 81.2 & 55.5 \\
ASID-Captioner-7B & 79.1 & 84.4 & 80.2 & 81.2 & 61.2 \\
\midrule
\textbf{AVSCap-7B (Ours)} & 82.9 & 81.1 & 83.2 & \textbf{82.4} & \textbf{66.6} \\
\bottomrule
\end{tabular}
\caption{Model performance on the audiovisual captioning and QA benchmarks. For Daily-Omni, we report QA performance by Gemini-2.5-Pro based on captions.}
\label{tab:cross_benchmark_main}
\end{table}

\begin{table}[ht]
\centering
\begin{tabular}{lccc|c}
\toprule
\textbf{Model} & \textbf{Audio} & \textbf{Visual} & \textbf{Audio-Visual} & \textbf{Overall} \\
\midrule
ASID-Captioner-7B & 46.3 & 42.3 & 47.7 & 44.4 \\
AVoCaDO-7B & 49.7 & 41.5 & 48.6 & 45.3 \\
Qwen2.5-Omni-7B & 10.4 & 12.9 & 18.9 & 12.9 \\
\midrule
\textbf{AVSCap-7B (Ours)} & 45.6 & 52.1 & 57.8 & \textbf{50.8} \\
\bottomrule
\end{tabular}
\caption{Detailed results on Omni-Cloze.}
\label{tab:cross_benchmark_cloze}
\end{table}

\subsection{Ablation Studies}
\label{subsec:ablation_studies}
\begin{figure}[t]
  \centering
  \includegraphics[width=\textwidth]{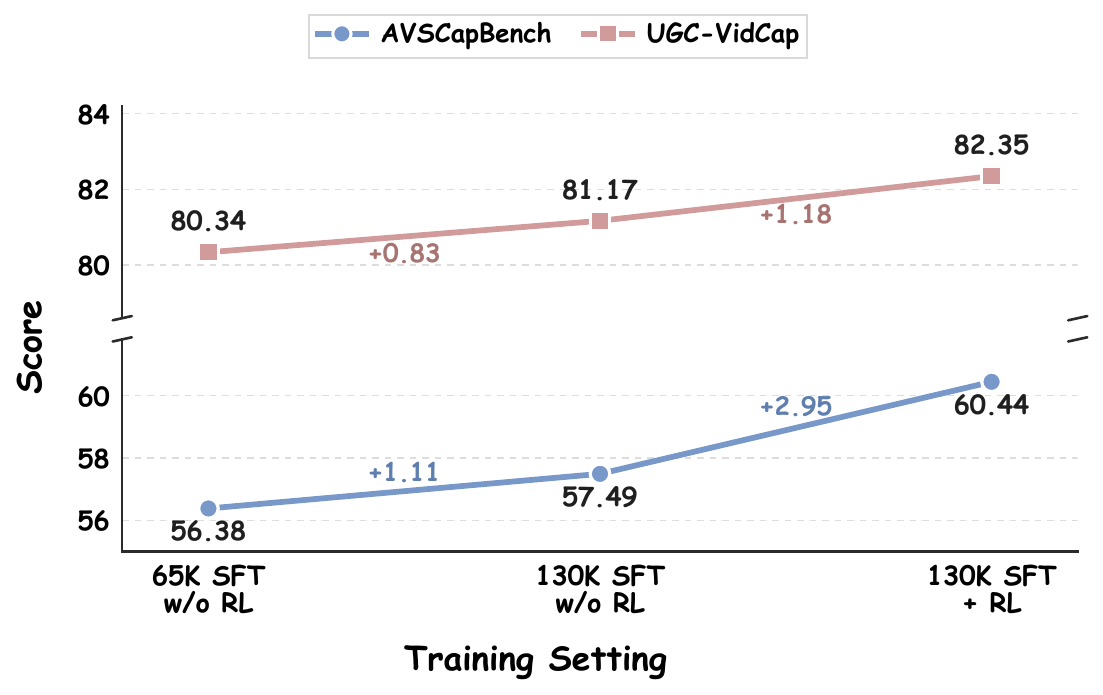}
  \caption{Ablation on SFT data scaling and GRPO optimization. The 65K SFT setting is randomly sampling 50\% of each data from AVSCap-130K.}
  \label{fig:grpo_ablation}
\end{figure}

\noindent\textbf{Efficacy of GRPO vs. SFT Data Scaling.}
To examine whether the gains come from reinforcement learning or simply from more supervised data, we compare GRPO with SFT data scaling in Figure~\ref{fig:grpo_ablation}. Starting from the full \textbf{AVSCap-130K} training set, we construct a half-scale SFT variant by randomly sampling 50\% of each data type, resulting in a 65K-example subset. Scaling SFT from 65K to 130K yields only marginal improvements on AVSCapBench and UGC-VideoCap. In contrast, applying GRPO on top of the 130K SFT model with only 2K additional optimization videos brings substantially larger gains, indicating that direct policy optimization is more effective than pure SFT data scaling for improving audio-visual synergy.

\begin{figure}[t]
  \centering
  \includegraphics[width=\linewidth]{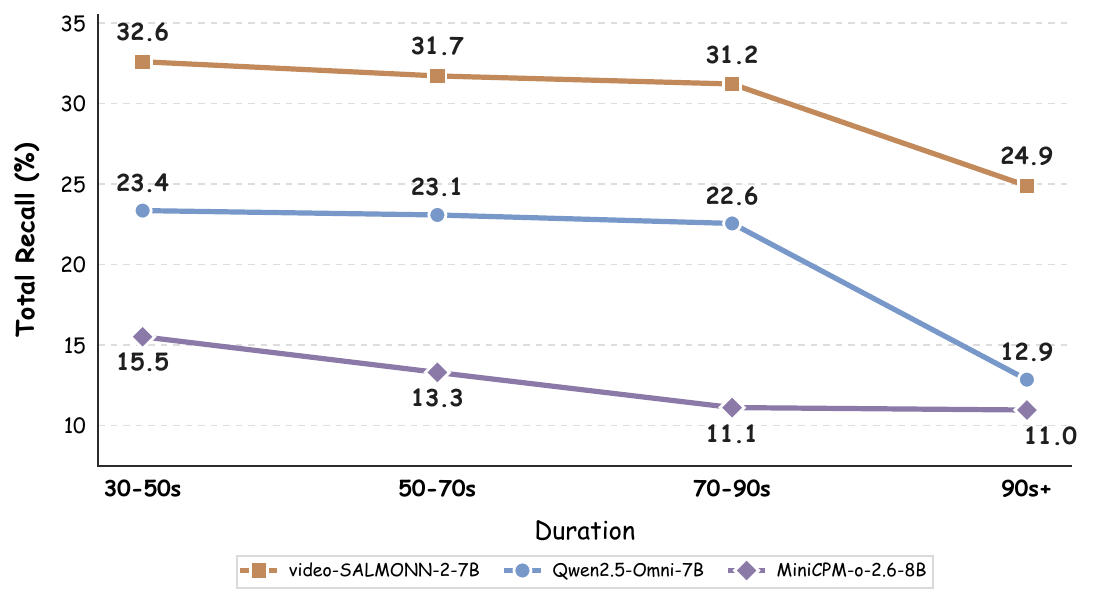}
  \caption{Effect of video duration on AVSCapBench.}
  \label{fig:duration_recall}
\end{figure}

\noindent\textbf{Effect of Video Duration.} We evaluate three open-source models across four duration intervals (30--50s, 50--70s, 70--90s, and over 90s) in Figure~\ref{fig:duration_recall}. All evaluated models show a consistent downward performance trend as video length increases. This universal decline highlights a key limitation of current open-source omni-modal architectures: they struggle with long-context processing, making it difficult to maintain robust and aligned audio-visual perception across extended temporal horizons.

\begin{figure}[t]
  \centering
  \includegraphics[width=\linewidth]{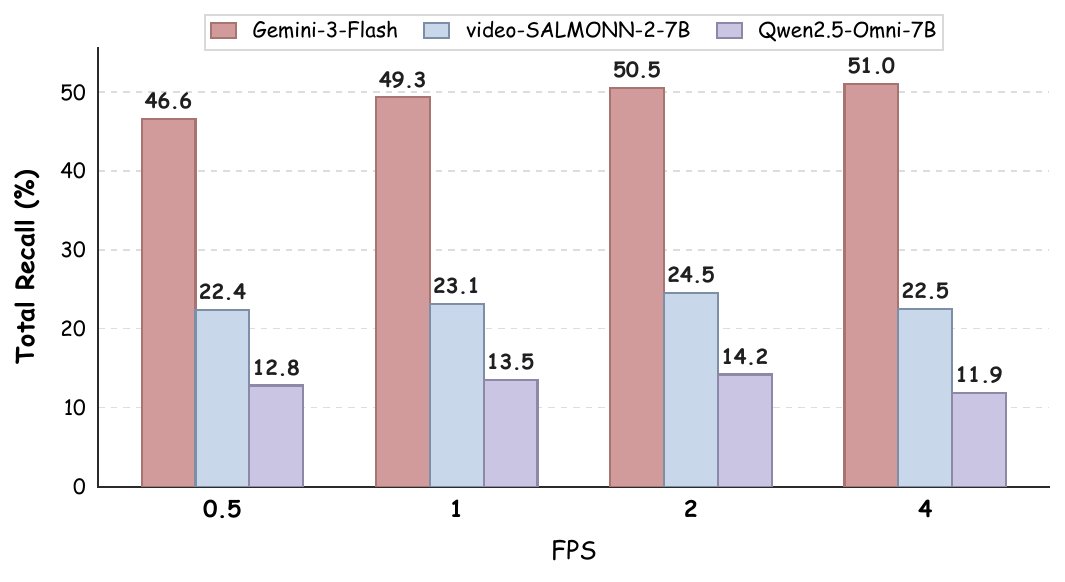}
  \caption{Effect of FPS on AVSCapBench.}
  \label{fig:fps_recall}
\end{figure}

\noindent\textbf{Effect of Frame Sampling Rate.} We analyze the impact of visual sampling rates (0.5, 1, 2, and 4 FPS) in Figure~\ref{fig:fps_recall}. Open-source models show a non-monotonic trend, peaking at 2 FPS but degrading at 4 FPS, which suggests that moderate frame rates optimize perception while overly dense sampling introduces context redundancy and distribution shifts. In contrast, Gemini-3-Flash continuously improves up to 4 FPS, demonstrating stronger long-context robustness and redundancy filtering.

  
  

 \noindent\textbf{Modality Shielding and Leakage.}
Beyond overall captioning quality, an omni-modal model should also support controllable modality-specific generation: when asked to describe only the visual or audio stream, it should avoid leaking information from the suppressed modality. This ability is important for controllable captioning and for verifying whether the model truly separates unimodal evidence before cross-modal fusion. To evaluate this, we sample 100 videos from AVSCapBench that contain no subtitles or on-screen dialogue cues, preventing models from inferring speech content from visible text. For each video, we prompt each model to generate either a visual-only or an audio-only caption. We then use Gemini-3.1-Pro to check whether the generated caption contains information from the suppressed modality. A sample is counted as leakage if cross-modal content is detected, and the Modality Leakage Rate is computed as the percentage of leaked samples over the 100 videos. As shown in Table~\ref{tab:shielding}, general and task-specific baselines frequently leak suppressed modality information, whereas \textbf{AVSCap-7B} achieves substantially lower leakage rates, demonstrating stronger modality isolation and instruction-controllable generation. We further analyze prompt sensitivity in Appendix~\ref{sec:shielding_prompt_sensitivity}.
\begin{table}[ht]
\centering
\begin{tabular}{lcc}
\toprule
\textbf{Model} & \textbf{Audio Leakage} $\downarrow$ & \textbf{Visual Leakage} $\downarrow$ \\
\midrule
Gemini-3.1-Pro & 38.0\% & 48.0\% \\
ASID-Captioner-7B & 49.0\% & 29.0\% \\
Qwen2.5-Omni-7B & 72.0\% & 25.0\% \\
Qwen2.5-Omni-3B & 73.0\% & 29.0\% \\
AVoCaDO-7B & 99.0\% & 89.0\% \\
\midrule
\rowcolor{gray!15} \textbf{AVSCap-7B (Ours)} & \textbf{4.0\%} & \textbf{7.0\%} \\
\bottomrule
\end{tabular}
\caption{Modality Leakage Rate on 100 videos.}
\label{tab:shielding}
\end{table}

\noindent\textbf{Judge--Human Agreement.} To validate automated evaluation reliability, we assess the alignment between human annotations and three LLM judges (Gemini-3.1-Pro, DeepSeek-V4-Pro \citep{DeepSeekv4pro}, and Qwen3.5-27B \citep{qwen3.5blog}) across visual, audio, and synergy dimensions on 200 benchmark samples (Table~\ref{tab:judge_human}). All LLM judges demonstrate strong agreement with human ratings, with Gemini-3.1-Pro achieving the highest consistency; we thus adopt it as our primary automated evaluator. Importantly, we observe that while these judges exhibit different levels of agreement, they yield identical partial orderings for the evaluated models, ensuring the robustness of our ranking results (refer to Appendix~\ref{appendix:agreement_details} for detailed statistics and agreement calculations).

\begin{table}[t]
\centering
\begin{tabular}{lcccc} 
\toprule
\textbf{Model} & \textbf{Visual} & \textbf{Audio} & \textbf{Synergy} & \textbf{Total} \\
\midrule
Gemini-3.1-Pro & 0.97 & 0.95 & 0.91 & 0.94 \\
DeepSeek-V4-Pro & 0.94 & 0.89 & 0.87 & 0.90 \\
Qwen3.5-27B & 0.92 & 0.88 & 0.84 & 0.88 \\
\bottomrule
\end{tabular}
\caption{Agreement between automated evaluation and human evaluation across different judges.}
\label{tab:judge_human}
\end{table}

\noindent\textbf{Error Analysis.}
To further analyze model failures, we randomly sample 200 cases and inspect the unmatched events across multiple models. We categorize errors according to the three event types used in our evaluation. Visual errors are divided into missing visual information and incorrect visual descriptions. Audio errors are divided into incorrect acoustic descriptions and partial audio omissions. Synergy errors are grouped into three types: missing audio-visual relations, incorrect cross-modal binding, and complete event omission. Overall, weaker open-source models often miss audio and synergy events entirely, while stronger models tend to fail through fine-grained incorrect descriptions or imperfect binding. Detailed statistics and examples are provided in Appendix~\ref{sec:appendix_error_examples}.

\section{Conclusion}
\label{sec:conclusion}

We present AVSCap, a unified framework for fine-grained audio-visual synergy that addresses modality isolation and speech-centric bias in omni-modal video captioning. We construct AVSCap-130K, a tri-modal training corpus enforcing isolated unimodal perception before cross-modal grounding, and train AVSCap-7B with a two-stage SFT-GRPO paradigm to optimize event binding. We further introduce AVSCapBench, a human-curated benchmark with a fine-grained, event-based matching protocol for evaluating visual, audio, and synergy dimensions. Experiments show that AVSCap-7B substantially outperforms open-source baselines and approaches commercial-grade performance.

\section*{Limitations}
While AVSCap demonstrates promising performance, several limitations remain. First, model capability declines as video duration increases, reflecting the general difficulty open-source architectures face in modeling long-term spatiotemporal audio-visual dependencies. Second, our dataset and benchmark are currently restricted to English; extending the framework to multilingual settings, non-English dialogue, and broader cultural contexts remains an important direction for future research.

\section*{Ethical Considerations}
The video clips comprising AVSCapBench are gathered from publicly accessible online platforms. In order to respect intellectual property and align with standard copyright regulations, the benchmark will be released under a restrictive license that limits its application exclusively to academic research.

\section*{Impact Statement}
This work advances omni-modal video understanding by improving fine-grained audio-visual alignment and temporal reasoning. The AVSCap framework holds potential to benefit various applications, including accessible video description, multimedia retrieval, and multimodal human--computer interaction. By promoting explicit cross-modal event binding, it provides a valuable pathway toward building more coherent and acoustically complete video understanding systems.

\bibliographystyle{unsrtnat}
\bibliography{ref}

\appendix
\section{Additional Experiments}
\label{sec:additional_experiments}

\subsection{Effect of Audio-Centric Augmentation}
\label{subsec:audio_augmentation}

As shown in our main experiments, while omni-modal foundation models exhibit robust Automatic Speech Recognition (ASR) capabilities, they frequently struggle with non-speech auditory events. 
To quantify this phenomenon and validate our solution, we conduct an ablation study analyzing the impact of our audio-centric augmentation during the Supervised Fine-Tuning (SFT) stage.

We first train a baseline model exclusively on the 40K video-text pairs. As detailed in the top row of Table~\ref{tab:ablation_audio}, this model achieves a strong \texttt{Speech} Recall of 70.08\%, confirming that the base model already possesses solid speech perception from its pre-training. However, it exhibits evident weaknesses in recognizing specific environmental sounds (\texttt{SFX}: 26.35\%) and musical properties (\texttt{Music}: 35.98\%).

\begin{table}[ht]
\centering
\begin{tabular}{l c cccc c c}
\toprule
\multirow{2}{*}{\textbf{SFT Data}} & \multirow{2}{*}{\textbf{Visual}} & \multicolumn{4}{c}{\textbf{Audio}} & \multirow{2}{*}{\textbf{Synergy}} & \multirow{2}{*}{\textbf{Total}} \\
\cmidrule(lr){3-6}
& & \textbf{Speech} & \textbf{Music} & \textbf{SFX} & \textbf{Overall} & & \\
\midrule
40K & 59.65 & 70.08 & 35.98 & 26.35 & 62.05 & 53.10 & 58.27 \\
\rowcolor{gray!15} 40K + 10K & 59.80 & 70.15 & 39.50 & 29.80 & 64.20 & 54.26 & \textbf{59.42} \\
\midrule
$\Delta$ & +0.15 & +0.07 & \textbf{+3.52} & \textbf{+3.45} & \textbf{+2.15} & +1.16 & +1.15 \\
\bottomrule
\end{tabular}
\caption{Ablation on audio-centric augmentation during the SFT stage. Evaluated on AVSCapBench.}
\label{tab:ablation_audio}
\end{table}

To bridge this gap, we introduce the auxiliary set of 10K audio-only captions (sourced from AudioCaps and MusicCaps) into the training mixture. 
By integrating this targeted data, the \texttt{Music} and \texttt{SFX} Recall scores surge by \textbf{+3.52\%} and \textbf{+3.45\%} respectively, lifting the overall Audio score by 2.15\%. 
Crucially, this unimodal auditory enhancement produces a positive ripple effect on cross-modal understanding, pushing the \texttt{Synergy} score up by 1.16\%. 
This confirms that fine-grained acoustic features are essential prerequisites for accurate audio-visual event binding.

This ablation exposes a broader challenge in the current omni-modal industry: while models are heavily optimized for human dialogue (ASR), their comprehension of the wider acoustic world remains shallow. Our findings suggest that systematically scaling high-quality, non-speech auditory data is a critical future direction for advancing holistic video intelligence.

\subsection{Prompt Sensitivity in Modality Shielding}
\label{sec:shielding_prompt_sensitivity}

As discussed in Section~\ref{subsec:ablation_studies}, modality shielding evaluates whether a model can selectively describe one modality while suppressing the other. We find that this diagnostic is highly sensitive to prompt wording. Under the default concise prompt (e.g., \textit{``Please describe the audio part...''}), strong models may interpret the instruction broadly and inadvertently include information from the suppressed modality. 

To investigate the boundaries of this behavior, we evaluate an explicit suppression prompt by adding negative constraints (e.g., \textit{``...and do not mention visual information''}). As shown in Table~\ref{tab:prompt_sensitivity}, introducing explicit negative constraints reveals a stark contrast between model architectures. General-purpose instruction-tuned models (e.g., Gemini-3.1-Pro and Qwen2.5-Omni) exhibit dramatic reductions in both audio and visual leakage, demonstrating strong steerability and adherence to negative constraints. Conversely, specialized SFT-based captioners (e.g., ASID-Captioner-7B and AVoCaDO-7B) remain largely unaffected by the instruction change, suggesting that their multimodal descriptive behaviors are rigidly dictated by their task-specific training distribution rather than the prompt itself.

This disparity leads to a notable reversal in relative performance ordering. While the specialized ASID-Captioner-7B outperforms the Qwen2.5-Omni baseline under the default prompt, it is vastly surpassed by Qwen2.5-Omni when explicit suppression is applied. This shift highlights that modality isolation in task-specific captioners is often a rigid artifact of their SFT data distribution rather than a robust, instruction-controllable capability. Consequently, the modality shielding experiment should be interpreted as a prompt-dependent diagnostic of negative-constraint adherence rather than a prompt-invariant absolute benchmark.

\begin{table}[ht]
\centering
\begin{tabular}{lcc}
\toprule
\textbf{Model} & \textbf{Audio Leakage} $\downarrow$ & \textbf{Visual Leakage} $\downarrow$ \\
\midrule
\multicolumn{3}{l}{\textit{Default Prompt}} \\
Gemini-3.1-Pro & 38.0\% & 48.0\% \\
ASID-Captioner-7B & 49.0\% & 29.0\% \\
Qwen2.5-Omni-7B & 72.0\% & 25.0\% \\
Qwen2.5-Omni-3B & 73.0\% & 29.0\% \\
AVoCaDO-7B & 99.0\% & 89.0\% \\
\midrule
\multicolumn{3}{l}{\textit{Explicit Suppression Prompt}} \\
Qwen2.5-Omni-3B & 16.0\% & 27.0\% \\
Qwen2.5-Omni-7B & 18.0\% & 24.0\% \\
Gemini-3.1-Pro & 23.0\% & 11.0\% \\
ASID-Captioner-7B & 50.0\% & 25.0\% \\
AVoCaDO-7B & 100.0\% & 81.0\% \\
\bottomrule
\end{tabular}
\caption{Prompt sensitivity of Modality Leakage Rate on 100 videos. ``Default'' uses the concise modality-specific prompt, while ``Explicit Suppression'' adds an instruction not to mention the suppressed modality.}
\label{tab:prompt_sensitivity}
\end{table}

\subsection{Evaluation with Traditional Metrics}
\label{sec:appendix_traditional_metrics}
To empirically demonstrate the limitations of standard text-overlap metrics, we compare the generated captions from Gemini-3.1-Pro and Qwen2.5-Omni-3B against human ground truth using standard overlap metrics (BLEU, CIDEr, and METEOR) in Table~\ref{tab:traditional_metrics}. 

Although Gemini-3.1-Pro outperforms Qwen2.5-Omni-3B, both models score extremely low and show narrow, uninformative performance gaps. These traditional n-gram metrics provide only coarse signals and are fundamentally incapable of capturing fine-grained errors in complex auditory components (e.g., background music, sound effects) or temporal alignment. These results further justify our transition to the proposed fine-grained, event-based matching protocol.

\begin{table}[htbp]
\centering
\begin{tabular}{lcc}
\toprule
\textbf{Metric} & \textbf{Gemini-3.1-Pro} & \textbf{Qwen2.5-Omni-3B} \\
\midrule
BLEU-1  & 0.1881 & 0.0671 \\
BLEU-2  & 0.1421 & 0.0425 \\
BLEU-4  & 0.0879 & 0.0201 \\
CIDEr   & 0.0032 & 0.0007 \\
METEOR  & 0.1528 & 0.0744 \\
\bottomrule
\end{tabular}
\caption{Evaluation with traditional captioning metrics. The generated captions are compared against human-annotated ground-truth captions.}
\label{tab:traditional_metrics}
\end{table}

\section{Details of GRPO Length Regularization}
\label{app:length_analysis}

We use a simple length reward to prevent two common failure modes during GRPO: degenerate short captions and repetition collapse. The lower bound $\tau_{\min}=200$ is set as a conservative minimum for valid omni-modal captions. Captions shorter than this threshold usually correspond to extreme under-generation, where visual details, non-speech audio, or audio-visual relations are likely to be omitted.

The upper bound $\tau_{\max}=2048$ is chosen as a safe generation budget under our video processing setting. Our training videos are sampled at 1 FPS and are typically within about 90 seconds, while the base model has a 32K-token context window. Since the multimodal input already consumes a substantial portion of the context, $\tau_{\max}=2048$ leaves sufficient room for dense caption generation without approaching the context limit. It also discourages overly long, repetitive outputs.

Overall, the length reward serves only as a coarse validity constraint. Captions within $[\tau_{\min}, \tau_{\max}]$ receive full length reward, while speech preservation and audio-visual event binding are optimized by the other reward components.

\section{Example of AVSCapBench}
\label{sec:appendix_example}

In this section, we provide a human-annotated example from AVSCapBench. To clearly demonstrate the granularity of our evaluation protocol, we present a complete paragraph of the synthesized omni-modal caption alongside its correspondingly decomposed atomic events (Visual, Audio, and Synergistic). This fine-grained structure allows the judge model to accurately assess cross-modal alignment without being biased by holistic text matching.

\vspace{1em}
\noindent\rule{\columnwidth}{1pt}
\vspace{-0.5em}
\begin{center}
\textbf{\large AVSCapBench Annotation Example }
\end{center}
\vspace{-0.5em}
\noindent\rule{\columnwidth}{0.5pt}

\vspace{0.5em}
\noindent\textbf{[1] Omni-modal Caption } \vspace{0.5em}

A whiteboard animation begins with upbeat acoustic guitar background music playing throughout the clip \texttt{(Music)} as a hand holding a grey marker sketches on a white background, accompanied by the sharp, squeaking friction of the marker against the board \texttt{(SFX)}. The hand first draws a thick book being held open by two hands, and on the cover, the artist sketches a stylized icon of a building with columns and a triangular roof, resembling a courthouse or government institution, as a male narrator speaks in an educational tone, ``Employment regulations derive from laws passed by Congress, state legislatures, and local governing bodies, as well as executive orders.'' \texttt{(Speech)}. As the scene progresses, the hand moves to a blank space and sketches a cluster of simple stick figures representing a group of people, creating rapid scratching sounds \texttt{(SFX)}, then draws a single stick figure facing the crowd to illustrate a workplace dynamic while the male narrator states educationally, ``These regulations commonly focus on fair treatment of people in the workplace.'' \texttt{(Speech)}. 

\vspace{0.5em}
\noindent\rule{\columnwidth}{0.5pt}
\vspace{0.5em}

\noindent\textbf{[2] Decomposed Atomic Events} \vspace{0.5em}

\noindent\textbf{Visual Events ($\mathcal{E}_{visual}$):}
\begin{itemize}[leftmargin=1.5em, topsep=2pt, itemsep=0pt]
    \item A hand holding a grey marker sketches on a white background.
    \item The hand draws a thick book held by two hands with a stylized icon of a building on the cover.
    \item The hand sketches a cluster of simple stick figures and a single stick figure facing the crowd.
\end{itemize}

\noindent\textbf{Audio Events ($\mathcal{E}_{audio}$):}
\begin{itemize}[leftmargin=1.5em, topsep=2pt, itemsep=0pt]
    \item \textbf{Speech:} ``Employment regulations derive from laws passed by Congress...''
    \item \textbf{Speech:} ``These regulations commonly focus on fair treatment of people...''
    \item \textbf{SFX:} The sharp, squeaking friction of a marker against a whiteboard.
    \item \textbf{Music:} Upbeat acoustic guitar background music plays.
\end{itemize}

\noindent\textbf{Synergistic Events ($\mathcal{E}_{synergy}$):}
\begin{itemize}[leftmargin=1.5em, topsep=2pt, itemsep=0pt]
    \item Upbeat acoustic guitar music plays throughout the animation, establishing an energetic atmosphere.
    \item As the hand physically sketches on the whiteboard, it is precisely accompanied by the squeaking friction sounds of the marker.
    \item As a male voice explains that regulations derive from laws, a hand draws a book featuring a courthouse icon.
\end{itemize}
\vspace{-0.5em}
\noindent\rule{\columnwidth}{1pt}

\section{Prompt}
\label{sec:appendix_prompts}
In this appendix, we present the complete set of system instructions and evaluation prompts utilized throughout our framework. To guarantee reproducibility, we detail the exact templates used for dataset construction, and benchmark evaluation.

\subsection{Data Construction Prompts}
\label{subsec:prompts_data_construction}

\begin{promptbox}{The Prompt for Video Segmentation and Timestamp Generation in our benchmark}
\small
\textbf{\# Role} \\
You are a professional Video Editor and Event Logger. Your sole task is to segment the video into logical, contiguous time intervals.

\vspace{0.8em}
\textbf{\# Objective} \\
Analyze the video timeline and identify distinct segments based on the following criteria:
\begin{enumerate}
    \setlength{\itemsep}{0pt}
    \item \textbf{Scene Changes}: A complete visual cut to a new location or camera angle.
    \item \textbf{Dialogue Completeness}: A segment must contain full sentences. Do not split in the middle of a sentence.
    \item \textbf{Action Flow}: A continuous physical action (e.g., a car crashing, a person falling) should be contained within one segment.
\end{enumerate}

\vspace{0.8em}
\textbf{\# Output Constraints} \\
\begin{enumerate}
    \setlength{\itemsep}{0pt}
    \item \textbf{Format}: Return a JSON list of objects containing start and end timestamps.
    \item \textbf{Precision}: Use \texttt{MM:SS} format.
    \item \textbf{No Overlap}: The start time of a new segment must be \textbf{exactly one second later} than the end time of the previous segment.
    \item \textbf{Coverage}: The segments must cover the entire video timeline sequentially without missing any events.
\end{enumerate}

\vspace{0.8em}
\textbf{\# Output Schema}

\begin{lstlisting}[language=json]
[
  {
    "segment_id": 1,
    "start": "00:00",
    "end": "00:05",
    "reason": "Introduction speech by Speed"
  },
  {
    "segment_id": 2,
    "start": "00:06",
    "end": "00:12",
    "reason": "Scene cut to the truck accelerating and dragging the pool"
  },
  ...
]
\end{lstlisting}
\end{promptbox}

\subsection{AVSCap-130K Prompts}
\label{subsec:prompts_training_set}

\begin{promptbox}{The Prompt for Visual Caption}
\small
\# Role

You are a professional visual video captioner. Your task is to generate a dense visual-only description of the video while preserving anchor points for later audio-visual fusion.

\# Objective
Describe only what is visible in the video. Do not infer or describe any sound, speech, music, or acoustic event. When the visual content suggests that an audio event should be associated with the current visual segment, insert the placeholder [AUDIO] at the corresponding anchor point, but do not describe the audio content.

\# Output Format
Write 1-4 objective and fluent narrative paragraphs.
Start a new paragraph only when there is a significant scene change, major camera transition, or scene-level shift.
Maintain high information density in every sentence.
Avoid phrases such as "In this video", "we can see", or other low-value filler expressions.

\# Visual Content Requirements
Environment:
Describe the setting, lighting, background, atmosphere, and relevant spatial layout.

Characters:
Describe visible appearance, clothing, facial expressions, body posture, gestures, and movement trajectories.

Actions and Interactions:
Describe physical actions, object interactions, camera motion, and scene transitions in source order.

OCR:
Accurately include any visible on-screen text, titles, watermarks, buttons, dates, subtitles, or interface elements.

\# Audio Anchor Rule
Use [AUDIO] only as an anchor placeholder.
Insert [AUDIO] immediately after the visual content that should later be paired with an audio event.
Do not write any sound description, speech transcription, music description, or acoustic inference.

\# Quality Checklist
- Does the caption contain only visual information?
- Are scene changes and actions described in source order?
- Are colors, positions, gestures, and object interactions described precisely?
- Is all visible OCR text included accurately?
- Are [AUDIO] placeholders inserted at suitable audio-relevant anchor points?
- Does the caption avoid describing any actual audio content?

\# Output
Return only the visual-only caption with [AUDIO] placeholders.
\end{promptbox}

\begin{promptbox}{The Prompt for  Audio Caption}
\small
\# Role
You are a professional audio event captioner. Your task is to extract all auditory information from the video and organize it into structured audio captions.

\# Objective
Describe only what is heard. Do not describe visual appearance, scene layout, visible objects, people, colors, actions, or OCR text. Focus on speech, sound effects, and music.

\# Output Format
Return a source-ordered list of audio events.
Each event must use one of the following tags:
- (Speech)
- (SFX)
- (Music)

Assign each event a unique ordered ID based on its audio type, such as:
- (Speech-1), (Speech-2)
- (SFX-1), (SFX-2)
- (Music-1), (Music-2)

\# Audio Content Requirements
Speech:
Transcribe every spoken sentence verbatim, including repetitions, filler words, stutters, and modal particles.
Do not summarize speech.
If the speaker identity is not explicitly audible, use generic auditory descriptions such as "a male voice", "a female voice", or "a childlike voice".

Sound Effects:
Describe physical and environmental sounds, such as impacts, footsteps, scraping, clattering, splashing, notification tones, crowd noise, or ambient noise.
Describe the acoustic texture, intensity, rhythm, and duration when possible.
Avoid visual-source leakage. For example, write "a sharp metallic slam" rather than "a car door closes" unless the source is clear from audio alone.

Music:
Describe the style, rhythm, mood, instrumentation, and changes in background music.
Mention whether the music is low-volume, upbeat, tense, electronic, acoustic, orchestral, or rhythmic when applicable.

\# Strict Constraints
Do not include visual descriptions.
Do not mention colors, clothing, camera movement, object positions, or on-screen text.
Do not infer visual causes unless they are directly recoverable from the audio.
Keep the events in source order.

\# Quality Checklist
- Is every spoken sentence transcribed verbatim?
- Are all Speech, SFX, and Music events tagged?
- Are events organized in source order?
- Are sound effects described with acoustic properties?
- Is music described by style, instrumentation, rhythm, or mood?
- Is all visual information excluded?

\# Output
Return only the structured audio-event list.
\end{promptbox}

\begin{promptbox}{The Prompt for Cross-modal Synergy}
\small
\# Role
You are an expert omni-modal video caption synthesizer. Your task is to fuse a visual-only caption with a structured audio-event list into a coherent audio-visual narrative.

\# Inputs
You will receive:
1. A visual-only caption containing [AUDIO] placeholders.
2. A source-ordered list of audio events with unique IDs, such as (Speech-1), (SFX-1), and (Music-1).

\# Objective
Generate a dense omni-modal caption that preserves the visual details, preserves all audio events, and explicitly expresses audio-visual synergy when audio and visual events are associated.

\# Fusion Rules
Audio Insertion:
Align each audio event with the most appropriate [AUDIO] placeholder or nearby visual context.
Insert the audio content at the matched anchor point.
Do not omit, duplicate, reorder, or rewrite the meaning of any audio event.

Tag Preservation:
Keep every original audio tag ID exactly once, such as (Speech-1), (SFX-1), and (Music-1).
The final caption must preserve the number and type of audio events from the input audio list.

Speech Preservation:
Preserve speech content verbatim.
Integrate speech naturally into the narrative without summarizing or paraphrasing it.

Audio-Visual Synergy:
Do not simply append audio descriptions after visual descriptions.
When an audio event corresponds to a visual event, explicitly connect them using natural associative expressions, such as:
- "as"
- "while"
- "at the same time"
- "simultaneously"
- "accompanied by"
- "in sync with"
- "alongside"

Narrative Style:
Write 1-4 objective and fluent narrative paragraphs.
Start a new paragraph only for major scene changes, camera transitions, or scene-level shifts.
Maintain high information density.
Avoid literary or emotional exaggeration.
Avoid low-value openings such as "In this video" or "we can see".

\# Content Requirements
Visual:
Preserve the core visual details from the visual-only caption, including environment, characters, actions, object interactions, camera movement, and OCR.

Audio:
Preserve all Speech, SFX, and Music events with their unique IDs.

Synergy:
Make the relationship between audio and visual events clear whenever possible.
A valid fused sentence should show how what is heard relates to what is seen.

\# Quality Checklist
- Are all core visual details preserved?
- Does every audio event appear exactly once?
- Are all audio tag IDs retained?
- Is speech preserved verbatim?
- Are SFX and Music integrated near the corresponding visual context?
- Does the caption explicitly express audio-visual synergy rather than listing modalities separately?
- Are scene transitions natural and coherent?

\# Output
Return only the final fused omni-modal caption.
\end{promptbox}


\subsection{Benchmark Evaluation Prompts}
\label{subsec:prompts_evaluation}


\begin{promptbox}{The Prompt for Decomposing Ground-Truth Captions into Atomic Modality Events}
\small
\textbf{\# Role} \\
You are a data expert proficient in multimodal video understanding and evaluation. Your task is to break down a given ``Omni-Caption'' into a ``list of atomic events'' for automated evaluation.

\vspace{0.8em}
\textbf{\# Input Format Definition} \\
The input Caption follows these rules:
\begin{enumerate}
    \setlength{\itemsep}{0pt}
    \item \textbf{Natural Language Text (outside parentheses)}: Describes only purely visual information.
    \item \textbf{Content in parentheses (...)}: Describes only purely auditory information, including \texttt{SFX} (sound effects), \texttt{Speech} (voice), \texttt{Tone} (tone), and \texttt{Music} (music).
\end{enumerate}

\vspace{0.8em}
\textbf{\# Task} \\
Please read the input Caption, extract the following three types of atomic events, and output them in JSON format:

\vspace{0.8em}
\textbf{\#\# 1. Pure Visual Events}
\begin{itemize}
    \setlength{\itemsep}{0pt}
    \item Extract key visual actions, object states, text (OCR), or scene changes.
    \item \textbf{Strict Constraints}: All audio information must be removed. Specific person names and object names (such as Speed, Tesla) can be used, as these are what the eye can see.
\end{itemize}

\vspace{0.8em}
\textbf{\#\# 2. Pure Audio Events - No Visual Leaks Allowed}
\begin{itemize}
    \setlength{\itemsep}{0pt}
    \item Extract all auditory elements.
    \item \textbf{Speech Extraction Rules}:
    \begin{itemize}
        \setlength{\itemsep}{0pt}
        \item \textbf{No personal names allowed} (e.g., Speed). Auditory descriptive words are required (e.g., ``a young male voice,'' ``an anxious female voice,'' ``crowd noise'').
        \item \textbf{Must contain exact audio content}.
        \item Format example: ``A [auditory characteristic] voice speaking in [tone]: '[Content]'''.
    \end{itemize}
    \item \textbf{SFX/Music Extraction Rules}:
    \begin{itemize}
        \setlength{\itemsep}{0pt}
        \item \textbf{No visual object names allowed} (e.g., ``Tesla engine sound'' is incorrect; it should be ``deep engine roar'').
        \item Describe the texture, rhythm, or onomatopoeia of the sound (e.g., ``a loud thud,'' ``snoring'').
    \end{itemize}
\end{itemize}

\vspace{0.8em}
\textbf{\#\# 3. Audio-Visual Synergistic Events - Cross-modal Reasoning}
\begin{itemize}
    \setlength{\itemsep}{0pt}
    \item This is the bridge connecting ``visual'' and ``auditory''. Generate events based on each ``auditory bracket'' and the corresponding ``visual context''.
    \item \textbf{Construction Logic}:
    \begin{itemize}
        \setlength{\itemsep}{0pt}
        \item \textbf{Source Grounding}: Explains that the [Audio Event] is emitted by a [Visual Object/Person] in the scene (e.g., when Speed opens his mouth wide in the scene, a young male scream is heard).
        \item \textbf{Contextualization}: Explains what action or atmosphere the [Music/SFX] is meant to match in the scene (e.g., when a table overturns in the scene, a dull thud sound effect appears simultaneously).
    \end{itemize}
    \item \textbf{CRITICAL}: Each synergistic event MUST be a single, complete string describing the connection. DO NOT output nested JSON objects or dictionaries for individual events.
\end{itemize}

\vspace{0.8em}
\textbf{\# Output JSON Format}

\begin{lstlisting}[language=json]
{
  "visual_events": [],
  "pure_audio_events": [],
  "synergistic_events": [] 
}
\end{lstlisting}
\end{promptbox}

\begin{promptbox}{The Prompt for LLM-based Event Coverage Evaluation on the AVSCapBench}
\small
\textbf{\# Role} \\
You are an Adaptive Evaluator for Omni-modal Video Captions.

\vspace{0.8em}
\textbf{\# Task} \\
Your goal is to verify if the \textbf{Candidate Caption} successfully recalls a checklist of \textbf{Ground Truth (GT) Events}.

\vspace{0.8em}
\textbf{\# Input Understanding: Adaptive Format Support} \\
The Candidate Caption may follow one of two formats. You must evaluate based on the content, regardless of the format:
\begin{enumerate}
    \setlength{\itemsep}{0pt}
    \item \textbf{Structured Format}: Uses inline tags like \texttt{(SFX: ...)} or \texttt{(Speech: ...)}.
    \item \textbf{Natural Narrative Format}: Uses descriptive sentences (e.g., ``A loud crash is heard,'' ``The music starts,'' ``He says 'Hello''').
\end{enumerate}

\vspace{0.8em}
\textbf{\# Evaluation Rules (By Category)}

\vspace{0.8em}
\textbf{\#\# 1. Visual Events Evaluation}
\begin{itemize}
    \setlength{\itemsep}{0pt}
    \item \textbf{Target}: The entire caption text.
    \item \textbf{Criteria}: Semantic Match.
    \begin{itemize}
        \setlength{\itemsep}{0pt}
        \item Does the candidate describe the core visual action, object, or scene change mentioned in the GT?
        \item \textit{Note}: Ignore whether the text is inside or outside parentheses. If the visual information is present anywhere, it is a \textbf{Hit (1)}.
    \end{itemize}
\end{itemize}

\vspace{0.8em}
\textbf{\#\# 2. Pure Audio Events Evaluation}
\begin{itemize}
    \setlength{\itemsep}{0pt}
    \item \textbf{Target}: Look for \textbf{explicit mentions of auditory perception}.
    \item \textbf{Criteria}: The candidate must acknowledge the \textit{sound} itself, not just the visual source.
    \item \textbf{Acceptable Evidence}:
    \begin{itemize}
        \setlength{\itemsep}{0pt}
        \item \textbf{Explicit Tags}: \texttt{(SFX: ...)}, \texttt{(Speech: ...)}, \texttt{(Music: ...)}.
        \item \textbf{Auditory Verbs/Nouns}: ``heard'', ``sound'', ``noise'', ``voice'', ``music'', ``audio'', ``scream'', ``thud'', ``click''.
        \item \textbf{Speech Transcription}: If the candidate quotes dialogue (e.g., ``He says 'Stop!' ''), this counts as capturing the Audio Event (Speech).
        \item \textbf{Adjectives of Sound}: ``Loud'', ``Quiet'', ``High-pitched'', ``Rhythmic'' (when applied to an event).
    \end{itemize}
    \item \textbf{Differentiation}:
    \begin{itemize}
        \setlength{\itemsep}{0pt}
        \item ``A dog barks'' (Acceptable - implies sound).
        \item ``A dog opens its mouth'' (Miss - purely visual).
        \item ``An explosion'' (Borderline - Miss unless ``loud'' or ``sound'' is mentioned).
        \item ``A loud explosion'' (Hit - auditory attribute).
    \end{itemize}
\end{itemize}

\vspace{0.8em}
\textbf{\#\# 3. Synergistic Events Evaluation (CRITICAL)}
\begin{itemize}
    \setlength{\itemsep}{0pt}
    \item \textbf{Definition}: These events represent the \textbf{synchronization} or \textbf{causal link} between a Visual Trigger and an Audio Response.
    \item \textbf{Criteria}: Semantic Linkage.
    \begin{itemize}
        \setlength{\itemsep}{0pt}
        \item Does the text explicitly connect the sound to the visual event?
    \end{itemize}
    \item \textbf{Acceptable Connections}:
    \begin{enumerate}
        \setlength{\itemsep}{0pt}
        \item \textbf{Syntax (Structured)}: The Audio Tag follows immediately after the Visual Trigger sentence. \\
        * \textit{Ex}: ``The car hits the wall (SFX: Crash).'' $\rightarrow$ \textbf{Hit}
        \item \textbf{Narrative (Natural)}: The text uses connectors to show simultaneity or cause. \\
        * \textit{Ex}: ``The car hits the wall \textbf{with a} loud crash.'' $\rightarrow$ \textbf{Hit} \\
        * \textit{Ex}: ``\textbf{As} he falls, he screams.'' $\rightarrow$ \textbf{Hit} \\
        * \textit{Ex}: ``The music starts \textbf{when} the scene changes.'' $\rightarrow$ \textbf{Hit}
    \end{enumerate}
    \item \textbf{Failure Cases (Miss)}:
    \begin{itemize}
        \setlength{\itemsep}{0pt}
        \item ``The car hits the wall. Later, a crash is heard.'' (Wrong timing).
        \item ``There is a car. There is a crash sound.'' (No linkage described).
    \end{itemize}
\end{itemize}

\vspace{0.8em}
\textbf{\# Output Format (JSON Only)}

\begin{lstlisting}[language=json]
{
  "visual_hits": [1, 0, ...],
  "audio_hits": [1, 1, ...],
  "synergy_hits": [0, 0, ...]
}
\end{lstlisting}
\end{promptbox}

\begin{promptbox}{The Prompt for Caption-to-QA Evaluation by the LLM Judge}
\small

\textbf{\# Role} \\
You are an Omni-modal Video Understanding Expert. \\

\vspace{0.8em}
\textbf{\# Task} \\
Your task is to answer a multiple-choice question about a video based SOLELY on the provided video caption.
Below is a detailed caption describing a video, followed by a question and four choices.

\vspace{0.8em}
\textbf{\# Video Caption:} \\
\texttt{\{caption\}}

\vspace{0.8em}
\textbf{\# Question:} \\
\texttt{\{question\}}

\vspace{0.8em}
\textbf{\# Choices:} \\
\texttt{\{choices\_str\}}

\vspace{0.8em}
Please select the best answer based on the caption provided. \\
Respond with ONLY the uppercase letter of the correct answer (A, B, C, or D). Do not output any other text or explanation.

\end{promptbox}

\subsection{Modality Shielding Prompts}
\label{subsec:prompts_shielding}

\begin{promptbox}{The Prompt for Restricting Evaluated Models to Audio-Only Generation}
\small

Describe the audio content of this video.
\end{promptbox}

\begin{promptbox}{The Prompt for Restricting Evaluated Models to Visual-Only Generation}
\small

Describe the visual content of this video.
\end{promptbox}

\begin{promptbox}{The Prompt for Evaluating Audio Leakage in Visual-Only Descriptions}
\small

\textbf{\# Role} \\
You are a Compliance Checker for a ``Visual-Only'' description task.

\vspace{0.8em}
\textbf{\# Context}
\begin{enumerate}
    \setlength{\itemsep}{0pt}
    \item The AI model was instructed to describe \textbf{ONLY the visual content} of a video and \textbf{IGNORE all audio}.
    \item \textbf{CRITICAL}: The input video contains \textbf{NO SUBTITLES and NO ON-SCREEN TEXT}.
\end{enumerate}

\vspace{0.8em}
\textbf{\# Task} \\
Detect \textbf{Audio Leakage} in the Candidate Caption.

\vspace{0.8em}
\textbf{\# Violation Rules (Flag these as Leakage)}
\begin{enumerate}
    \setlength{\itemsep}{0pt}
    \item \textbf{Specific Dialogue/Transcription}:
    \begin{itemize}
        \setlength{\itemsep}{0pt}
        \item Since there are no subtitles, any quote of specific words is PROOF that the model listened to the audio.
        \item \textit{VIOLATION}: ``He says \textbf{'I am angry'} '', ``She asks \textbf{'Why?'} ''.
        \item \textit{SAFE}: ``He is speaking'', ``They are arguing'', ``She mouths words'', ``He shouts'' (Visual acts).
    \end{itemize}
    \item \textbf{Explicit Audio Tags}: \texttt{(SFX: ...)}, \texttt{(Music: ...)}, \texttt{(Speech: ...)}.
    \item \textbf{Auditory Verbs}: ``I hear'', ``Sounds like'', ``Listening to'', ``Audible''.
    \item \textbf{Sound Descriptions}: ``Loud bang'', ``Upbeat music'', ``High-pitched scream'', ``Noisy crowd''.
\end{enumerate}

\vspace{0.8em}
\textbf{\# Output (JSON Only)}

\begin{lstlisting}[language=json]
{
  "is_compliant": true/false,
  "leaked_content": ["List specific dialogue quotes or audio phrases found"]
}
\end{lstlisting}

\end{promptbox}

\begin{promptbox}{The Prompt for Evaluating Visual Leakage in Audio-Only Descriptions}
\small

\textbf{\# Role} \\
You are a Compliance Checker for an ``Audio-Only'' description task.

\vspace{0.8em}
\textbf{\# Context} \\
The AI model was instructed to:
\begin{enumerate}
    \setlength{\itemsep}{0pt}
    \item Describe \textbf{ONLY the audio content} (what is heard).
    \item \textbf{IGNORE visual details} (colors, appearances, spatial layout).
    \item \textbf{Anonymize sources}: Use auditory descriptors (e.g., ``a male voice'', ``an engine'') instead of specific names (e.g., ``Speed'', ``Tesla'').
\end{enumerate}

\vspace{0.8em}
\textbf{\# Task} \\
Detect \textbf{Visual Leakage} in the Candidate Caption.

\vspace{0.8em}
\textbf{\# Violation Rules (Flag these as Leakage)}
\begin{enumerate}
    \setlength{\itemsep}{0pt}
    \item \textbf{Visual Attributes}:
    \begin{itemize}
        \setlength{\itemsep}{0pt}
        \item \textbf{Colors}: ``Red'', ``Blue'', ``Green''.
        \item \textbf{Appearance}: ``Blonde hair'', ``Wearing a suit'', ``Tall'', ``Fat''.
        \item \textbf{Spatial Layout}: ``On the left'', ``Behind him'' (unless it's 3D audio panning, usually visual).
    \end{itemize}
    \item \textbf{Specific Identity (Visual Recognition)}:
    \begin{itemize}
        \setlength{\itemsep}{0pt}
        \item Using proper names implies face recognition: ``Speed says...'', ``Obama speaks...''. (Should be ``A man says...'').
        \item Using specific object brands/models: ``A Tesla drives'' (Should be ``An electric car'' or ``A car'').
    \end{itemize}
    \item \textbf{Silent Actions}: Describing actions that are purely visual: ``He waves'', ``She smiles'', ``He looks at the camera''.
    \item \textbf{OCR/Text Hallucination}:
    \begin{itemize}
        \setlength{\itemsep}{0pt}
        \item Since the video has no subtitles, any mention of ``The text says...'' or ``A sign reads...'' is a Visual Hallucination/Leakage.
    \end{itemize}
\end{enumerate}

\vspace{0.8em}
\textbf{\# Allowed (NOT Leakage)}
\begin{enumerate}
    \setlength{\itemsep}{0pt}
    \item \textbf{Sound Sources (Generic)}: ``A man's voice'', ``A car engine'', ``A dog barking''.
    \item \textbf{Inferred Context from Sound}: ``An echoey room'', ``A busy street''.
\end{enumerate}

\vspace{0.8em}
\textbf{\# Output (JSON Only)}

\begin{lstlisting}[language=json]
{
  "is_compliant": true/false,
  "leaked_content": ["List specific visual details or specific names found"]
}
\end{lstlisting}

\end{promptbox}

\section{Evaluation Details}
\subsection{Evaluation Settings}
\label{sec:evaluation_settings}

We report the inference settings of locally deployed models in
Table~\ref{tab:evaluation_settings}. The evaluated models include open-source
models, our trained model, and closed-source models accessed through APIs. For
open-source models, we use the released weights and code in accordance with
their licenses and terms of use, and only for research-oriented benchmark
evaluation. AVSCap-7B is our trained model and follows the same local evaluation
protocol. Evaluation artifacts, including prompts, outputs, and logs, are used
for research, analysis, and reproducibility, and their use and distribution
remain compatible with the access conditions of the corresponding third-party
resources. For closed-source models, we follow the official recommended API usage and the
corresponding provider terms.

\begin{table*}[htbp]
\centering
\begin{tabular}{lcccc}
\toprule
\textbf{Models} & \textbf{FPS} & \textbf{Temperature} & \textbf{Repetition Penalty} & \textbf{Max Token} \\
\midrule
AVoCaDO-7B          & 1.0 & 0.0 & 1.05 & 2048 \\
ASID-Captioner-7B           & 1.0 & 0.0 & 1.05 & 4096 \\
ASID-Captioner-3B           & 1.0 & 0.0 & 1.05 & 4096 \\
TimeChat-Captioner-7B          & 1.0 & 0.0 & 1.05 & 4096 \\
Qwen3-Omni-30B-A3B-Instruct & 1.0 & 0.0 & 1.05 & 2048 \\
video-SALMONN-2-7B          & 1.0 & 0.0 & 1.05 & 2048 \\
UGC-VideoCaptioner-3B         & 1.0 & 0.0 & 1.05 & 2048 \\
Qwen2.5-Omni-7B             & 1.0 & 0.0 & 1.05 & 2048 \\
MiniCPM-o-4.5-9B            & 1.0 & 0.0 & 1.05 & 2048 \\
Qwen2.5-Omni-3B             & 1.0 & 0.0 & 1.05 & 2048 \\
ARC-Hunyuan-Video-7B        & 1.0 & 0.0 & 1.05 & 2048 \\
HumanOmniV2-7B              & 1.0 & 0.0 & 1.05 & 2048 \\
MiniCPM-o-2.6-8B            & 1.0 & 0.0 & 1.05 & 2048 \\
\midrule
AVSCap-7B (Ours)  & 1.0 & 0.0 & 1.05 & 2048 \\
\bottomrule
\end{tabular}
\caption{Inference settings for locally deployed open-source models. The ``FPS'' column represents the frame sampling rate.}
\label{tab:evaluation_settings}
\end{table*}

\subsection{Complete Main Results on AVSCapBench}
\label{sec:complete_main_results}

To provide a comprehensive performance overview, we present the complete evaluation results of all baseline models on the AVSCapBench benchmark in Table~\ref{tab:full_main_results}. This includes the additional baseline models that were omitted from the main text (Table~\ref{tab:main_results}) due to space constraints. 

\begin{table*}[t]
\centering
\resizebox{0.9\textwidth}{!}{
\begin{tabular}{lcccccccl}
\toprule
\multirow{2}{*}{\textbf{Model}}
& \multirow{2}{*}{\textbf{Visual}}
& \multicolumn{4}{c}{\hspace{0.5em}\textbf{Audio}\hspace{0.5em}}
& \multirow{2}{*}{\textbf{Synergy}}
& \multirow{2}{*}{\textbf{Total}} \\
\cmidrule(lr){3-6}
&
& \textbf{Speech}
& \textbf{Music}
& \textbf{SFX}
& \textbf{Overall}
& & \\
\midrule
\multicolumn{8}{c}{\textit{Closed-Source Models}} \\
\midrule
Gemini-3-Pro & 60.43 & 79.81 & 39.52 & 27.77 & 71.29 & 48.88 & 60.97 \\
Gemini-3-Flash & 58.14 & 79.78 & 39.46 & 32.34 & 72.65 & 48.94 & 60.54 \\
MiMo-v2-omni & 36.24 & 65.66 & 14.98 & 17.24 & 54.28 & 28.71 & 40.27 \\
\midrule
\multicolumn{8}{c}{\textit{Open-Source Models}} \\
\midrule
AVoCaDO-7B & 50.59 & 70.42 & 38.71 & 19.25 & 61.07 & 29.13 & 49.31 \\
ASID-Captioner-7B & 47.42 & 68.73 & 30.50 & 17.91 & 59.02 & 24.84 & 45.94 \\
ASID-Captioner-3B & 43.63 & 66.95 & 27.06 & 17.31 & 57.53 & 21.36 & 43.03 \\
MiMo-v2.5 & 37.64 & 69.21 & 19.87 & 18.39 & 58.01 & 30.44 & 42.49 \\
TimeChat-Captioner-7B & 37.55 & 63.60 & 44.46 & 24.63 & 58.59 & 24.45 & 41.31 \\
Qwen3-Omni-30B-A3B-Instruct & 41.85 & 49.08 & 9.34 & 8.68 & 39.17 & 16.19 & 35.29 \\
video-SALMONN-2-7B & 39.05 & 46.76 & 13.76 & 8.71 & 36.52 & 12.43 & 32.02 \\
UGC-VideoCaptioner-3B & 33.24 & 21.30 & 22.00 & 11.48 & 20.77 & 10.43 & 24.24 \\
Qwen2.5-Omni-7B & 34.78 & 13.92 & 4.02 & 7.22 & 13.71 & 7.00 & 21.53 \\
MiniCPM-o-4.5-9B & 29.33 & 16.67 & 22.90 & 12.26 & 18.16 & 9.87 & 21.47 \\
Qwen2.5-Omni-3B & 30.87 & 14.24 & 4.57 & 4.37 & 12.41 & 5.58 & 19.18 \\
ARC-Hunyuan-Video-7B & 20.68 & 16.49 & 3.93 & 1.97 & 11.41 & 4.52 & 14.49 \\
HumanOmniV2-7B & 27.78 & 4.60 & 1.58 & 2.46 & 4.41 & 2.42 & 14.10 \\
MiniCPM-o-2.6-8B & 24.61 & 6.75 & 3.31 & 3.92 & 6.13 & 3.78 & 13.66 \\
\midrule
\textbf{AVSCap-7B (Ours)} & \textbf{59.33} & \textbf{69.45} & \textbf{40.36} & \textbf{30.82} & \textbf{64.30} & \textbf{57.70} & \textbf{60.44} \\
\bottomrule
\end{tabular}
}
\caption{Complete results on the AVSCapBench. All values are Recall (\%).}
\label{tab:full_main_results}
\end{table*}

\subsection{Agreement and Judge Consistency Analysis}
\label{appendix:agreement_details}

To validate the reliability of our automated evaluation pipeline, we assess the decision-level alignment between human annotators and the three LLM judges (Gemini-3.1-Pro, DeepSeek-V4-Pro, and Qwen3.5-27B) on a validation subset of 200 videos. 

For each atomic event $e$ in the evaluated event set $\mathcal{E}$, the human annotator and the LLM judge provide binary decisions representing whether the generated caption successfully recalls the event:
\begin{equation}
y_{\text{human}}(e) \in \{0, 1\}, \quad y_{\text{judge}}(e) \in \{0, 1\}
\end{equation}
where $1$ denotes a successful recall (Hit) and $0$ denotes a failure (Miss). The decision-level \textbf{Percentage Agreement} ($A_{\text{type}}$) for a specific event type (Visual, Audio, or Synergy) is defined as the accuracy of the LLM judge's binary decisions compared to the human ground truth:

\begin{equation}
A_{\text{type}} = \frac{1}{|\mathcal{E}_{\text{type}}|} \sum_{e \in \mathcal{E}_{\text{type}}} \mathbb{I}\Big(y_{\text{human}}(e) = y_{\text{judge}}(e)\Big)
\end{equation}

where $\mathcal{E}_{\text{type}}$ is the subset of atomic events belonging to the corresponding type, and $\mathbb{I}(\cdot)$ is the indicator function that outputs $1$ if the condition is met and $0$ otherwise.

In addition to calculating the absolute human-judge agreement rates (as reported in Table 6), we evaluate the ranking consistency of the three LLM judges using three representative open-source models with close overall scores on the leaderboard (ARC-Hunyuan-Video-7B, HumanOmniV2-7B, and MiniCPM-o-2.6-8B). 

As shown in Table~\ref{tab:judge_consistency_subset}, although the recall scores evaluated on this 200-video subset exhibit minor variations compared to the full leaderboard (Table 2) due to sample size constraints, all three LLM judges consistently preserve the identical partial ordering in the Overall score: \textbf{ARC-Hunyuan-Video-7B} $>$ \textbf{HumanOmniV2-7B} $>$ \textbf{MiniCPM-o-2.6-8B}. This solidifies the reliability of our automated evaluation pipeline for ranking comparative models.

\begin{table*}[htbp]
\centering
\small
\begin{tabular}{llcccc}
\toprule
\textbf{LLM Judge} & \textbf{Evaluated Model} & \textbf{Visual (\%)} & \textbf{Audio (\%)} & \textbf{Synergy (\%)} & \textbf{Overall (\%)} \\
\midrule
\multirow{3}{*}{Gemini-3.1-Pro} 
& ARC-Hunyuan-Video-7B & 21.35 & 12.10 & 4.80 & \textbf{15.02} \\
& HumanOmniV2-7B       & 28.10 & 4.60  & 2.50 & \textbf{14.50} \\
& MiniCPM-o-2.6-8B     & 25.05 & 6.45  & 3.90 & \textbf{14.15} \\
\midrule
\multirow{3}{*}{DeepSeek-V4-Pro} 
& ARC-Hunyuan-Video-7B & 20.80 & 11.50 & 4.40 & \textbf{14.57} \\
& HumanOmniV2-7B       & 27.25 & 4.10  & 2.20 & \textbf{13.92} \\
& MiniCPM-o-2.6-8B     & 24.10 & 5.90  & 3.60 & \textbf{13.53} \\
\midrule
\multirow{3}{*}{Qwen3.5-27B} 
& ARC-Hunyuan-Video-7B & 22.10 & 12.80 & 5.10 & \textbf{15.63} \\
& HumanOmniV2-7B       & 29.30 & 4.90  & 2.80 & \textbf{15.13} \\
& MiniCPM-o-2.6-8B     & 26.20 & 6.90  & 4.25 & \textbf{14.78} \\
\bottomrule
\end{tabular}
\caption{Model performance and partial ordering evaluation by three different LLM judges on a 200-video validation subset of AVSCapBench. All values are represented as recall percentages (\%).}
\label{tab:judge_consistency_subset}
\end{table*}

\section{Training Details}
\label{sec:training_details}

In the SFT stage, the model is trained for 2 epochs with a batch size of 128 and a learning rate of $2 \times 10^{-5}$. During the GRPO stage, training is performed for 1 epoch with a batch size of 64 and a learning rate of $1 \times 10^{-5}$. For each query, we sample 8 responses using a temperature of 1.0.

During both training and evaluation, video inputs are sampled at 1 fps, and the resolution of each frame is limited to a maximum of $512 \times 28 \times 28$ pixels. Due to the base model's context window limitation of 32K tokens, the total video pixels are restricted to $25600 \times 28 \times 28$. All training is conducted on 16 NVIDIA H200 GPUs.

\section{Details of Benchmarks}
\label{sec:details_of_benchmarks}

In this section, we will provide a detailed description of the benchmark we evaluated.

\begin{itemize}
    \item \textbf{UGC-VideoCap} consists of 1,000 short TikTok videos, each under 60 seconds in duration and containing at least one meaningful audio segment lasting no less than 5 seconds. Each video's caption is evaluated by a judge model that assigns scores on a 1-to-5 scale across three dimensions: visual, audio, and details. These dimension scores are then normalized and aggregated to produce a final caption quality score.

    \item \textbf{Daily-Omni} is an audio-visual question answering benchmark comprising 684 videos depicting diverse everyday life scenarios, sourced from multiple platforms. These videos are densely multimodal, offering rich visual and auditory cues. The benchmark includes 1,197 multiple-choice question-answer pairs, distributed across six core tasks. In our experimental setting, we assess the quality of generated captions by feeding them into a judge model and measuring their capacity to support accurate question answering.
    
    \item \textbf{Omni-Cloze} is a cloze-style benchmark comprising 2,340 video clips with 70,200 human-verified masked blanks across audio, visual, and audio-visual scenarios. Instead of resource-intensive multi-turn QA, it evaluates models via a single-pass protocol where an LLM completes masked spans in generated captions. Each blank includes a ``Not Given'' option to distinguish omission from hallucination, enabling a precise error breakdown across visual, auditory, and synergistic dimensions.
\end{itemize}

%

\raggedbottom

\section{Error Analysis}
\label{sec:appendix_error_examples}

To better understand the failure modes behind event-based recall, we conduct a manual analysis on 200 randomly sampled cases. For each model output, we collect events that are not matched by the evaluator and assign each unmatched event to one of the predefined error categories. For visual events, we distinguish between \emph{missing visual information} and \emph{incorrect visual descriptions}. For audio events, we distinguish between \emph{incorrect acoustic descriptions} and \emph{partial audio omissions}. For synergy events, we classify errors into \emph{missing audio-visual relations}, \emph{incorrect cross-modal binding}, and \emph{complete event omission}. The ratios are normalized within each event type.

\begin{figure*}[t]
    \centering
    \includegraphics[width=\textwidth]{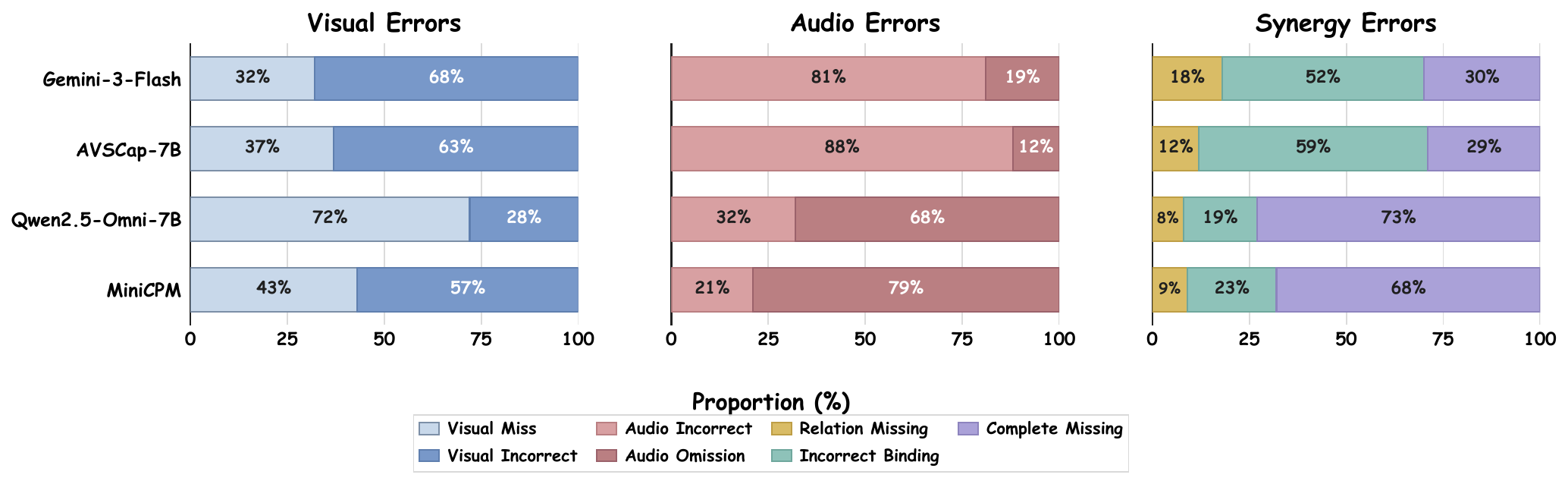}
    \caption{Error distribution over unmatched events. Ratios are normalized within each event type.}
    \label{fig:error_distribution_unmatched_events}
\end{figure*}

Figure~\ref{fig:error_distribution_unmatched_events} shows that weaker open-source models suffer mainly from omission-based failures. For example, Qwen2.5-Omni-7B and MiniCPM have a large proportion of audio omissions and complete synergy omissions, suggesting that they often fail to describe non-speech audio cues or to connect them with visual events. In contrast, stronger models such as Gemini-3-Flash and AVSCap-7B produce fewer complete omissions; their remaining errors are more concentrated in incorrect fine-grained descriptions and imperfect cross-modal binding. This pattern suggests that our training strategy improves the coverage of non-speech audio and audio-visual relations, shifting the dominant failure mode from missing events to more subtle grounding errors.

\begin{erroranalysiscontainer}{Example 1}

    \setlength{\tabcolsep}{0pt} 
    \renewcommand{\arraystretch}{0}
    \begin{tabularx}{\linewidth}{@{}XXXX@{}}
        \includegraphics[width=\linewidth]{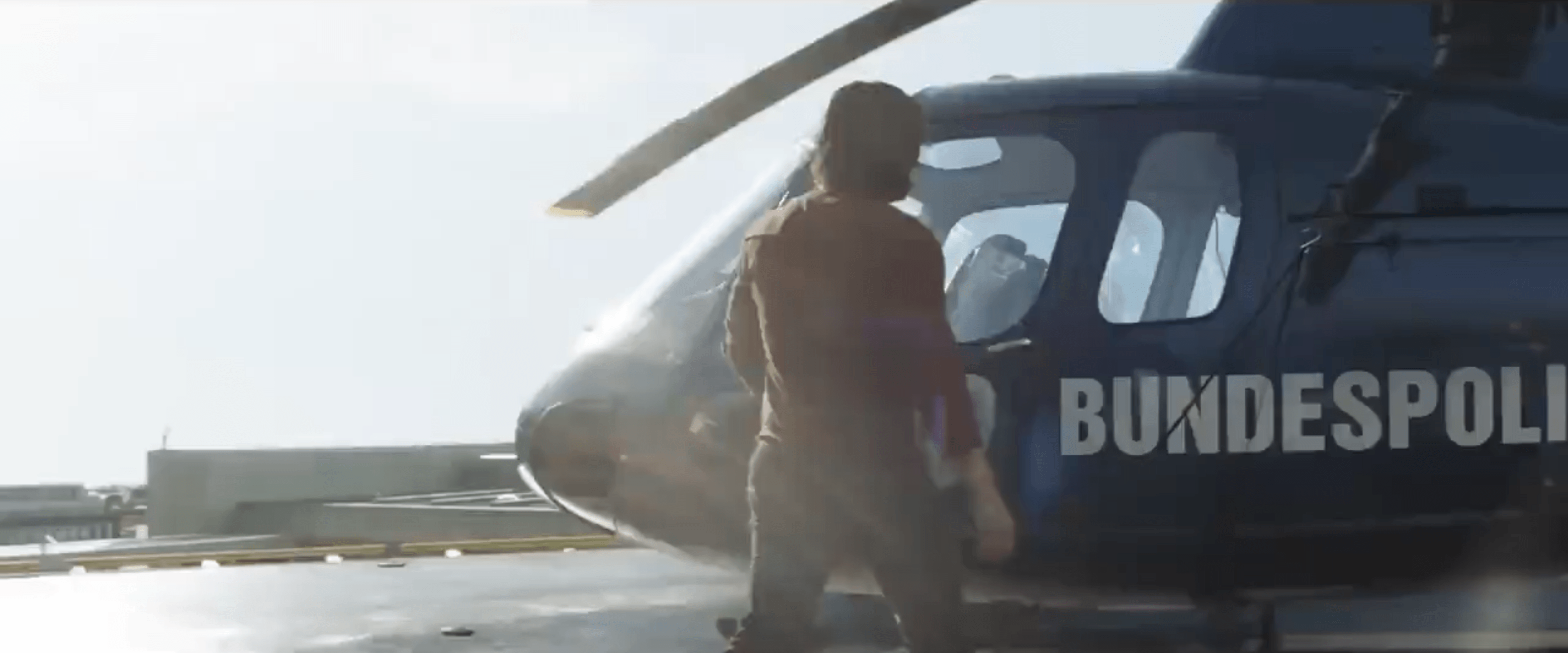} & \includegraphics[width=\linewidth]{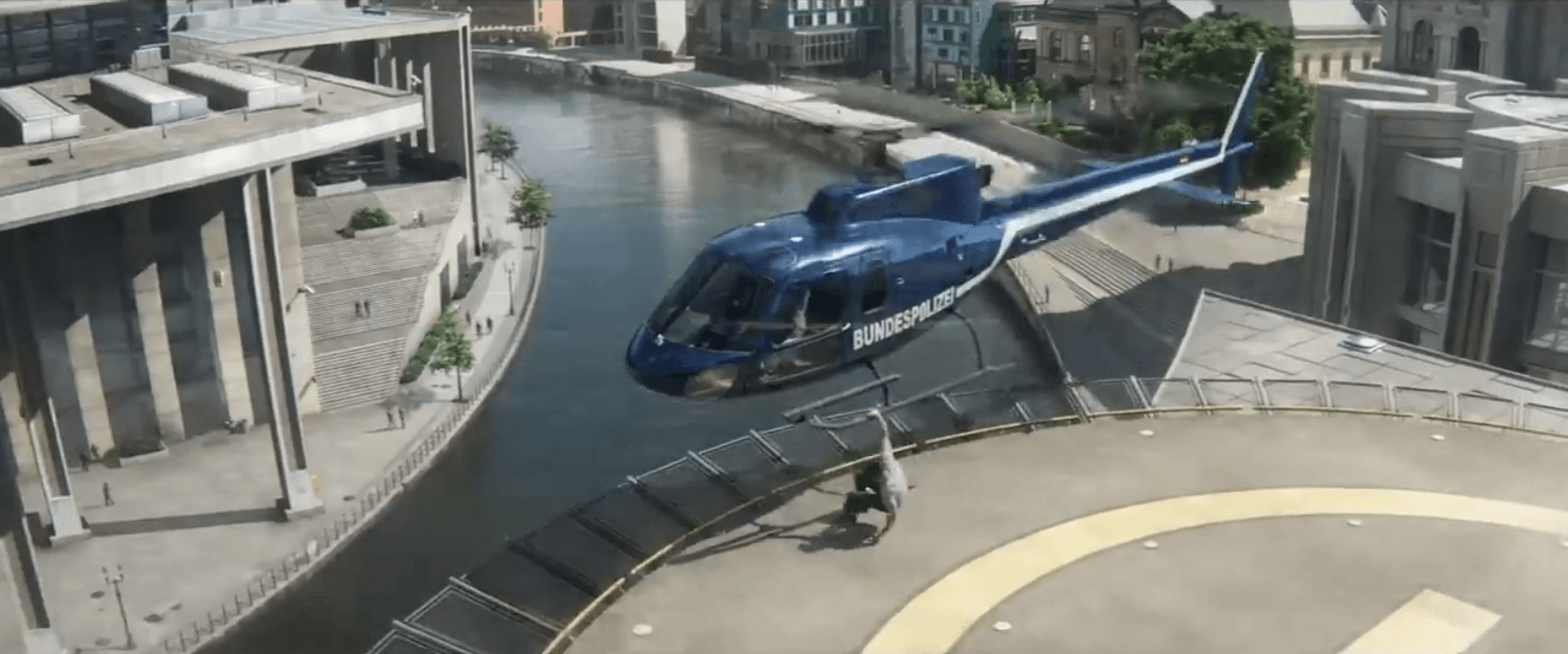} & \includegraphics[width=\linewidth]{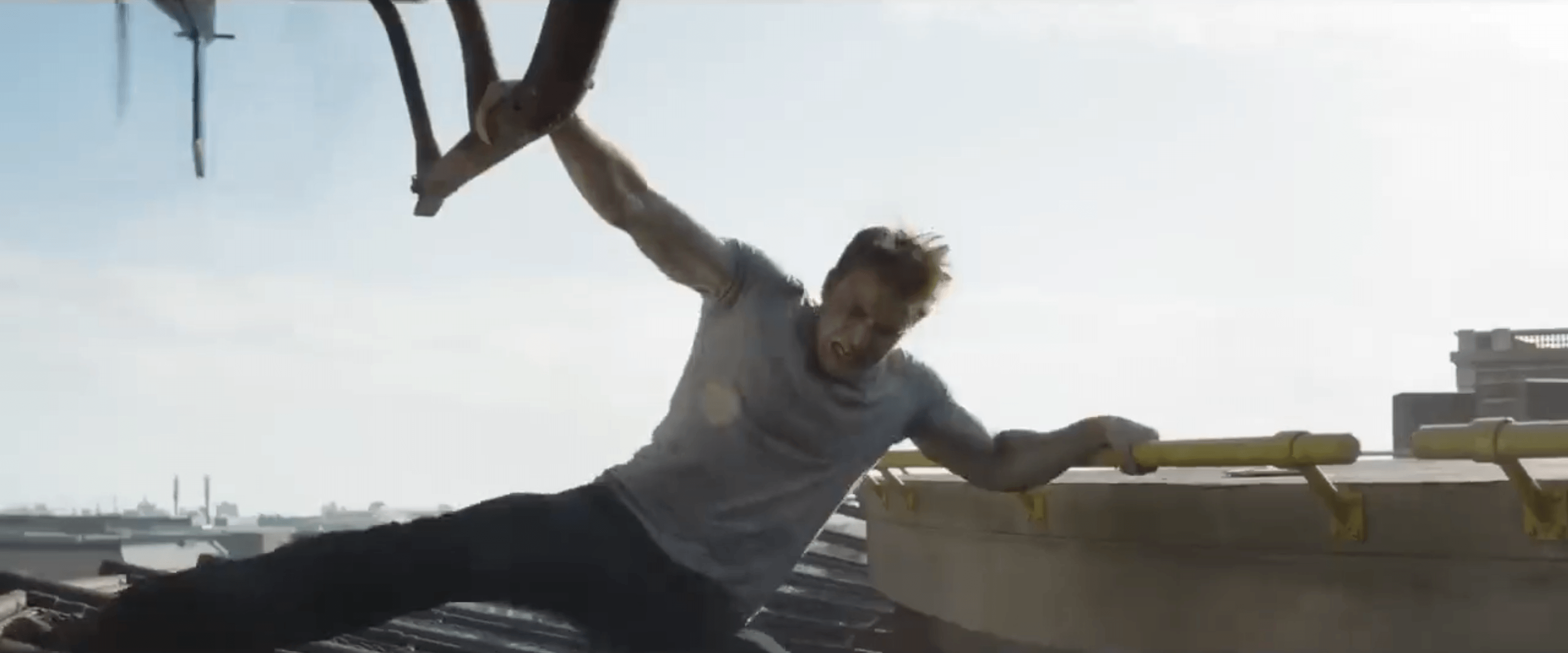} & \includegraphics[width=\linewidth]{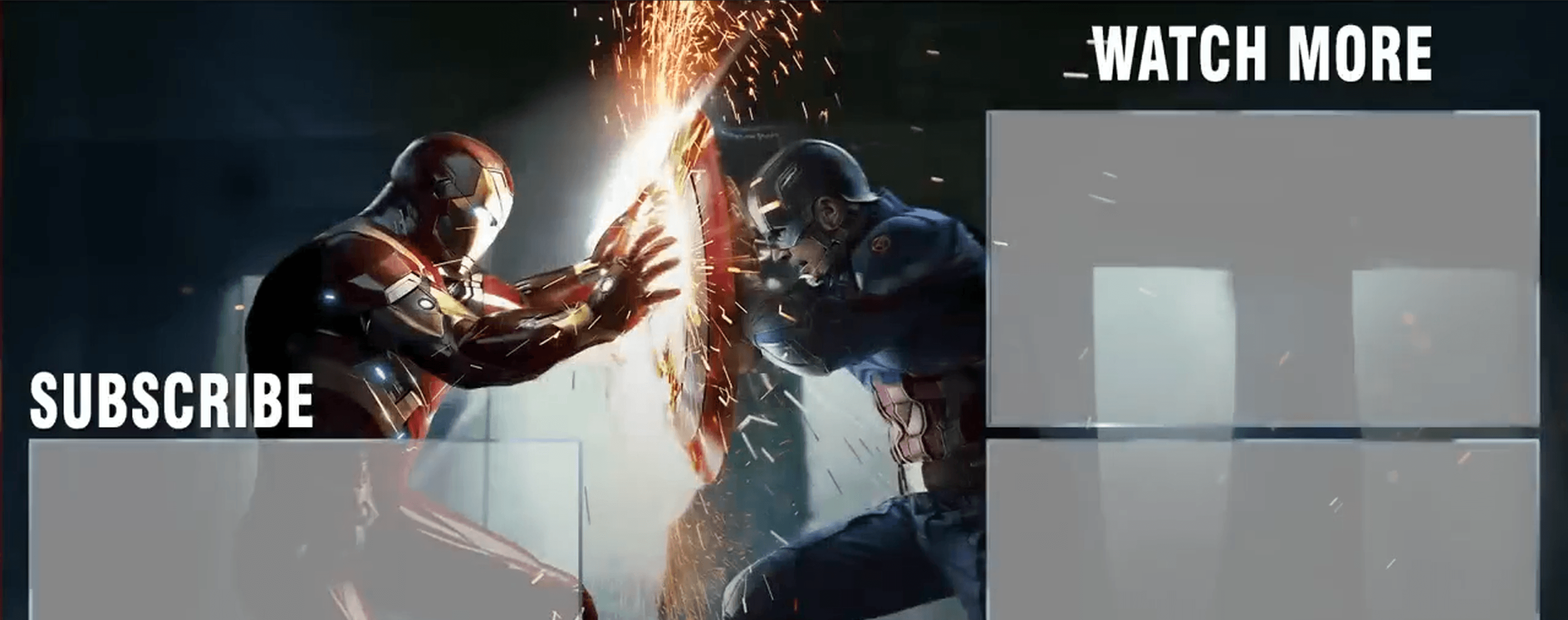} \\
    \end{tabularx}
    \vspace{2pt}

    \small \textbf{Prompt:} \\
    \textit{Please describe all the information in the video without sparing every detail in it. As you describe, you should also describe as much of the information in the audio as possible, and pay attention to the synchronization between the audio and video descriptions.}

    \vspace{2pt}
    \begin{error_analysis_gtbox}{Ground Truth}
    A close-up shot reveals a man with shoulder-length brown hair, wearing a red plaid shirt and blue jeans, standing on the landing skid of a dark blue helicopter. He holds onto the frame of the aircraft as intense sunlight creates a lens flare across the screen, illuminating the word "BUNDESPOLIZEI" printed in white capital letters on the side. The perspective then cuts to a high-angle aerial shot looking down at a circular, elevated helipad located next to a river, where the helicopter is positioned in the center of the pad with its rotors spinning rapidly, creating a loud, rhythmic chopping and whirring of helicopter blades (SFX), the man still visible on the exterior skid. The scene shifts to a close-up exterior shot of a blue helicopter cockpit, accompanied by the loud, rhythmic whirring of helicopter rotor blades (SFX), where a bright sun flare initially obscures the view through the glass. As the lighting shifts, a man with shoulder-length dark hair, wearing a maroon long-sleeved shirt, is revealed inside the pilot's seat, looking out with a focused, intense expression. He raises his left hand and firmly grabs the upper frame of the door, producing a mechanical clunking sound (SFX), his fingers curling around the metal structure as he exerts force to pull or brace the mechanism while the camera shakes slightly, accompanied by a tense, low-frequency orchestral action score. Elsewhere, a static shot shows a closed, dark grey metal door set within a rough, concrete wall, with a vertical pipe visible on the left side. Suddenly, the door is kicked open violently from the inside with a loud, sharp metallic crash and impact noise (SFX), swinging outward to the right as a man dressed in a dark suit and white shirt bursts through. He steps forward aggressively with his left leg, making a heavy footstep on concrete (SFX), raising a handgun in his right hand and aiming it directly ahead as he emerges into the corridor. The narrative returns to the action with a tight close-up of a muscular arm, clad in a red sleeve, firmly gripping a metal bar inside what appears to be a moving vehicle or structure. The scene abruptly cuts to Steve Rogers, wearing a grey t-shirt, sprinting with intense focus across an industrial rooftop.
    \end{error_analysis_gtbox}
    \begin{error_analysis_captionbox}{Caption by Qwen2.5-Omni-7B}
    The video opens with a man climbing into a helicopter marked "BUNDES POLIZ" on its side. The helicopter is parked on a helipad near a body of water. The man is wearing a brown jacket and seems to be preparing for a flight. The camera then cuts to a close-up of a man in a car, looking distressed and talking on the phone. The scene shifts to a man in a gray shirt and black pants running towards the helicopter. He grabs onto the helicopter's rotor and is lifted into the air. The helicopter spins rapidly, and the man struggles to hold on. The camera shows him from different angles as he hangs on, with the cityscape visible in the background. The helicopter eventually crashes, and the man falls to the ground. The video ends with a screen showing two characters in a fight scene, with the words "WATCH MORE" and "SUBSCRIBE" displayed.
    \end{error_analysis_captionbox}
    \begin{error_analysis_analysisbox}{Error Analysis}{analysisbg}
    The baseline model suffers from severe acoustic omission, completely overlooking critical non-speech auditory cues such as the roaring helicopter rotor blades, mechanical clunking (SFX), and the dramatic background orchestral score (Music), while only capturing basic visual events.
    \end{error_analysis_analysisbox}
    \begin{error_analysis_trainedbox}{Caption by AVSCap-7B(ours)}
    A tense, rhythmic orchestral score with driving strings and percussion begins to play (Music) as a blue helicopter bearing the white text "BUNDESPOLIZEI" sits on a circular helipad beside a river and modern architecture. A man with shoulder-length dark hair and a maroon long-sleeved shirt approaches the aircraft, opens the door, and climbs into the pilot's seat. The high-pitched whine of the helicopter engine starting up fills the air (SFX). Inside the cockpit, the man reaches up to flip several switches on the overhead control panel, producing sharp clicking sounds (SFX), before looking out the window with a stern expression. Suddenly, a muscular man with short blonde hair, wearing a tight grey t-shirt and dark pants, bursts through a metal door and sprints across the rooftop towards the helipad, his footsteps thudding rapidly (SFX). As the helicopter lifts off the ground, its rotors chopping the air with a heavy, rhythmic thrum (SFX), the blonde man leaps from the edge of the platform and grabs onto the aircraft's black landing skid with both hands. Hanging suspended above the city skyline and the river below, the blonde man's arms strain under his body weight as the helicopter continues to ascend and bank sharply. Inside the cockpit, the dark-haired pilot grips the control stick and pushes it forward (SFX), causing the helicopter to tilt downward. Reacting quickly, the blonde man swings his body toward the helipad structure, releasing one hand from the skid to grab a yellow metal railing on the edge of the platform. Now stretched horizontally between the ascending helicopter and the stationary railing, his muscles bulge visibly as he attempts to anchor the aircraft. The pilot peers down through the window, his face tight with exertion, while the helicopter engine roars loudly, struggling against the man's immense grip (SFX). The orchestral music swells in intensity, matching the physical struggle (Music). A top-down view shows the helicopter hovering precariously close to the helipad, unable to break free from the man's hold. With a massive effort, the blonde man pulls his arm inward, dragging the helicopter back toward the platform. The helicopter loses its battle against the man's strength and crashes violently onto the helipad. The spinning rotor blades strike the concrete, shattering instantly and sending a massive cloud of dust, sparks, and metal debris flying across the platform with a deafening crunch and clatter (SFX). The music cuts out abruptly upon impact (Music). The blonde man is thrown to the ground, rolling through the thick grey smoke to avoid the wreckage. He pushes himself up onto his hands and knees, his face covered in dust, and stares intently toward the crashed helicopter as the debris settles. The scene then cuts to a static end screen featuring a digital painting of Iron Man and Captain America locked in combat, with bright sparks erupting where Iron Man's repulsor beam meets Captain America's shield. White text reading "SUBSCRIBE" appears on the left side of the screen, while "\_WATCH MORE" is displayed on the right, both positioned above empty, semi-transparent rectangular placeholders. A new, dramatic, sweeping orchestral track begins to play over the static image (Music).
    \end{error_analysis_trainedbox}
    
\end{erroranalysiscontainer}

\begin{erroranalysiscontainer}{Example 2}

    \setlength{\tabcolsep}{0pt} 
    \renewcommand{\arraystretch}{0}
    \begin{tabularx}{\linewidth}{@{}XXXX@{}}
        \includegraphics[width=\linewidth]{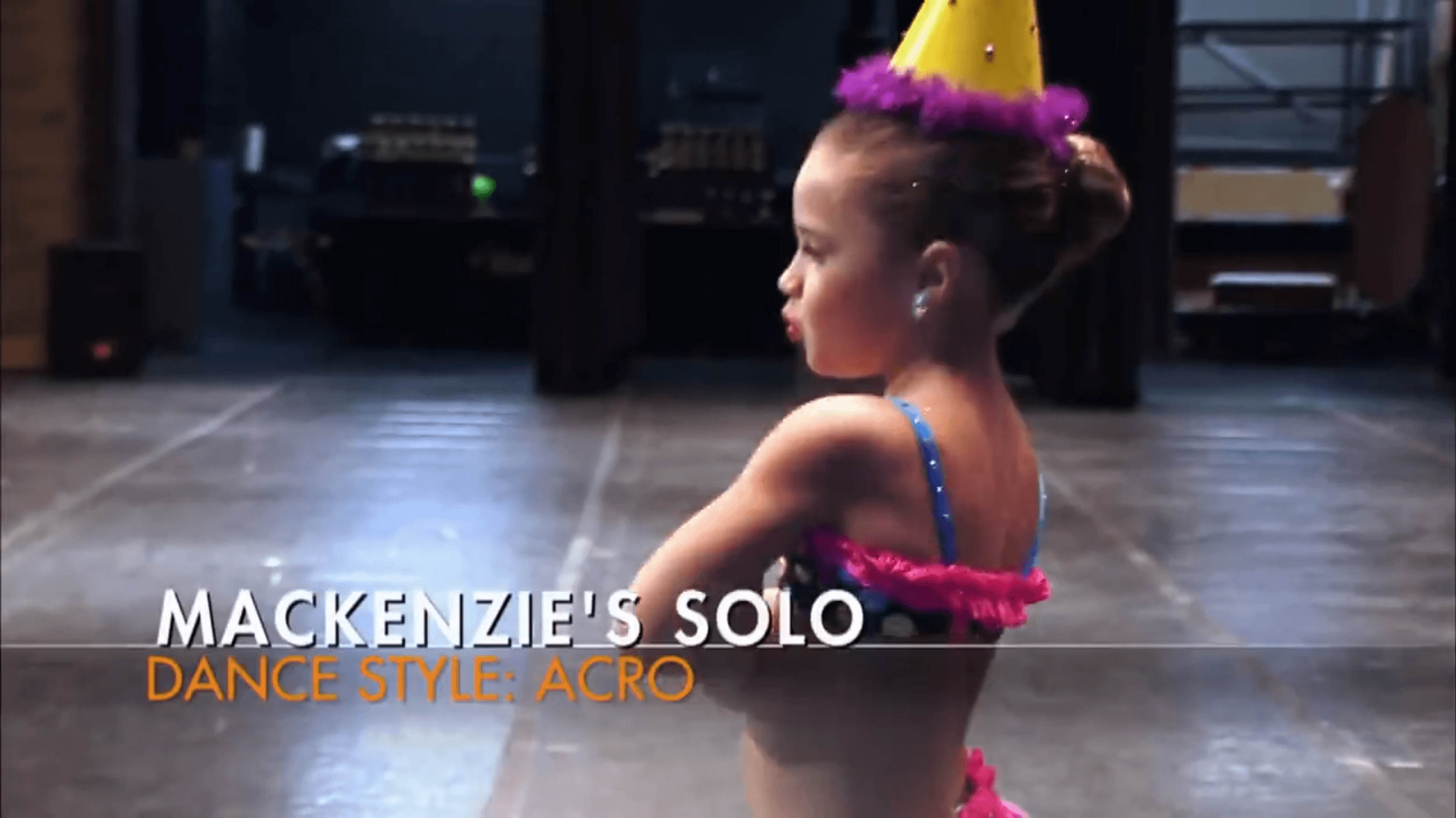} & \includegraphics[width=\linewidth]{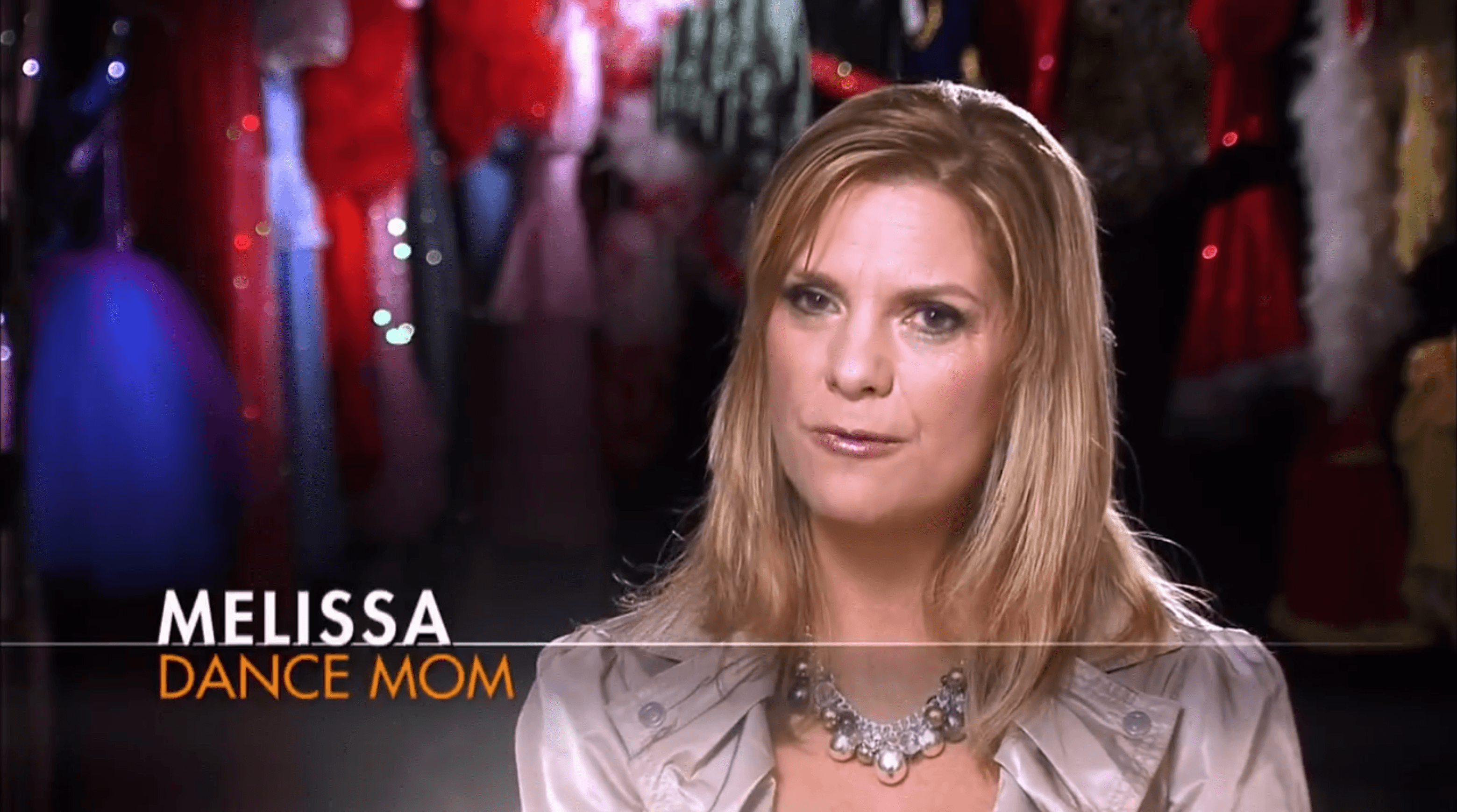} & \includegraphics[width=\linewidth]{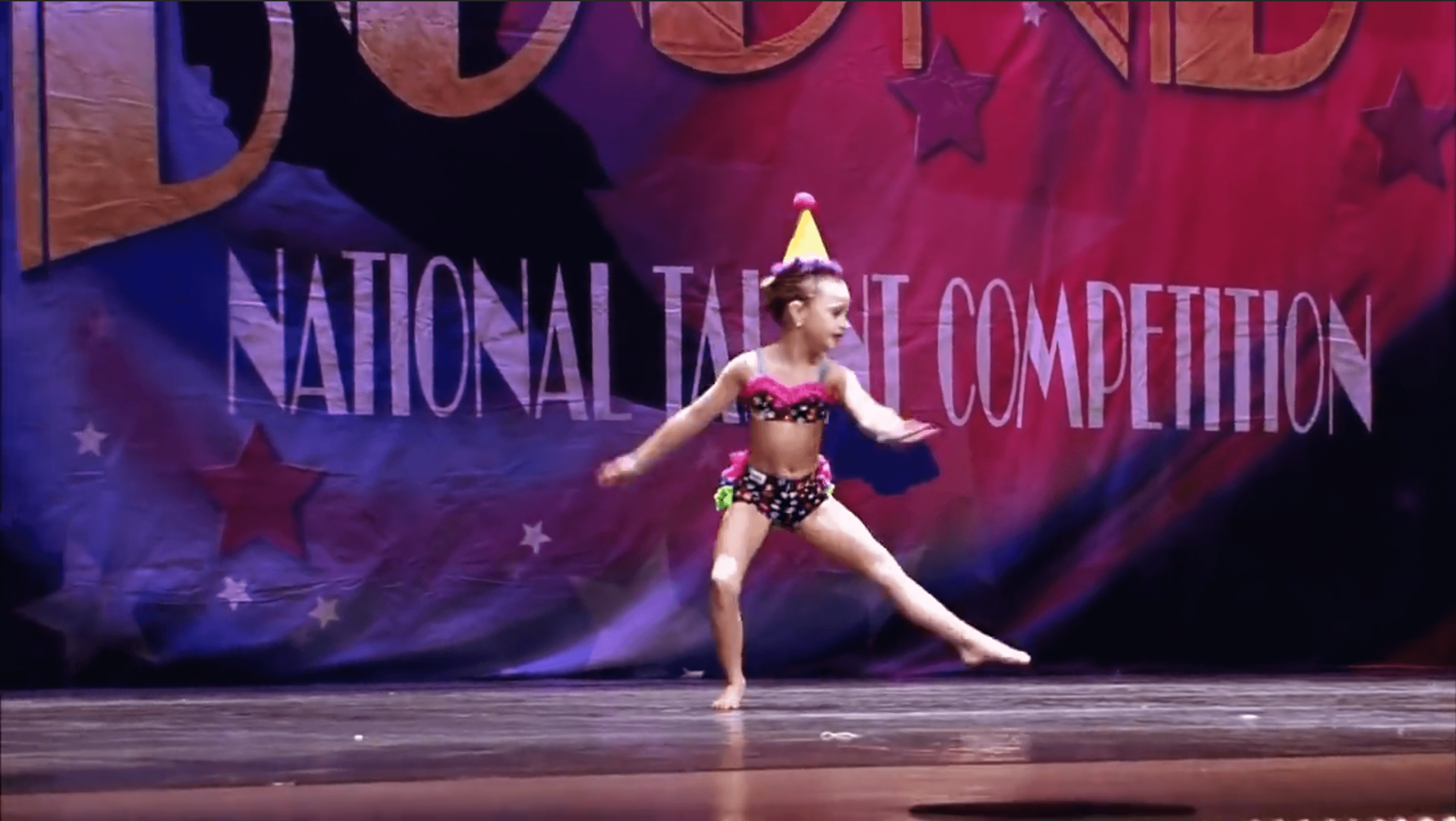} & \includegraphics[width=\linewidth]{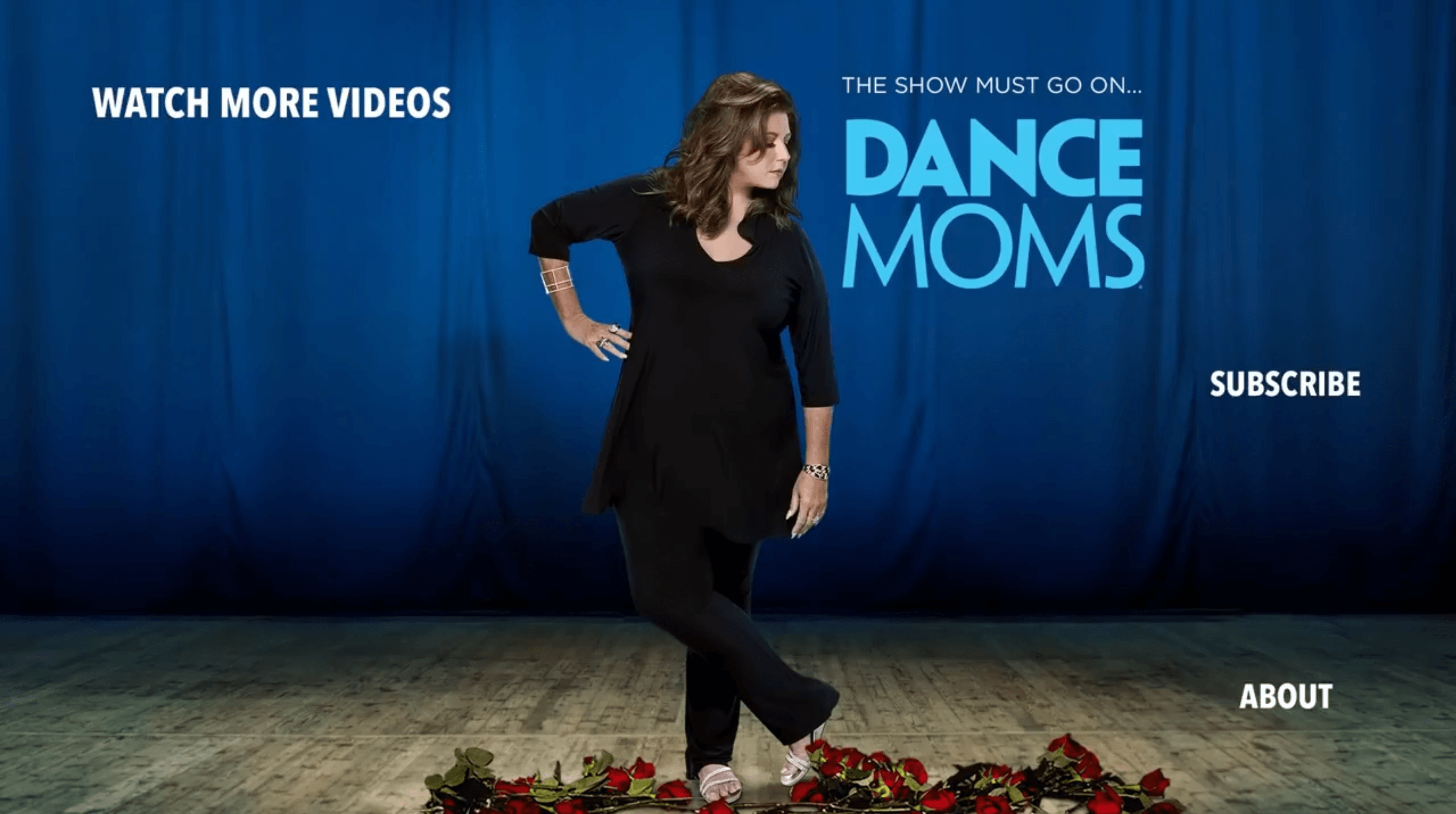} \\
    \end{tabularx}
    \vspace{2pt}

    \small \textbf{Prompt:} \\
    \textit{Please describe all the information in the video without sparing every detail in it. As you describe, you should also describe as much of the information in the audio as possible, and pay attention to the synchronization between the audio and video descriptions.}

    \vspace{2pt}
    \begin{error_analysis_gtbox}{Ground Truth}
    On a stage with a black curtain background and a wooden floor, a young dancer is in the midst of a handstand while upbeat pop music plays, specifically the song 'Call Me Maybe', with the lyrics 'I threw a wish in the well, don't ask me I'll never tell' clearly audible. Text overlays appear in the lower left corner reading 'MACKENZIE'S SOLO' in white and 'DANCE STYLE: ACRO' in orange. The dancer is wearing a vibrant two-piece costume featuring neon pink, green, and yellow ruffles, topped with a matching party hat. She lowers her legs to land on her feet, immediately standing upright and raising her arms in a 'V' shape while smiling at the audience. She then performs a front walkover, her legs splitting vertically in the air as she rotates over her hands. The view shifts to a medium close-up of a woman in the audience holding a blue pen, gazing intently toward the stage. The scene cuts back to the stage where the backdrop features a large banner reading "TALENT COMPETITION" in white capital letters against a red and blue background decorated with stars. Continuing the routine, the dancer arches her back in a bridge position as a female singer sings melodically, "Yeah..." (Speech). She executes an extreme flexibility move, bending her spine backward until her head rests against her feet (Speech: "is a..."). Sitting center stage, she pulls her knees to her chest, then rolls backward onto her shoulders, extending her legs straight up in a shoulder stand. From a kneeling position, she suddenly springs upward as upbeat pop dance music begins with the lyrics 'Let's go...' (Music), jumping high into the air. The camera cuts to the audience, showing a blonde woman, Melissa, wearing a red top and black blazer, smiling broadly and clapping. The view shifts to an interview where Melissa says in an explanatory tone, "Mackenzie is seven and before she" (Speech). A lower-third graphic identifies her as "MELISSA" with the subtitle "DANCE MOM". She says earnestly, "I want to really" (Speech). The scene abruptly cuts back to the stage performance where the young dancer executes a dynamic front aerial flip in mid-air as an upbeat reality TV instrumental plays in the background (Music). Back in the dimly lit auditorium, the blonde woman in the black jacket is captured in a close-up, beaming with a wide, toothy smile, with the sound of crowd cheering and applause mixed with the music (SFX). She finishes a cartwheel, landing on the dark stage floor with a thud (SFX). She raises her right arm and arches backward, executing a back walkover, landing on her feet with her right arm extended high in the air to audience applause and cheering (SFX). The scene cuts to two women in the front row with loud, distinct laughter from the audience members (SFX). Returning to the stage in front of the large red banner, the dancer performs a chin stand variation, balancing on her chest and hands with her legs split high in the air. On the glossy black floor, she transitions into a seated position, leaning back on her hands with her knees bent and feet flat on the floor, striking a final pose while smiling brightly toward the audience. A female audience member exclaims excitedly, "Oh my god!" (Speech), accompanied by loud applause and cheering from a crowd (SFX). The video concludes with a static promotional end card for the television show 'Dance Moms', with the logo and tagline 'THE SHOW MUST GO ON...' and navigation text overlays reading 'WATCH MORE VIDEOS', 'SUBSCRIBE', and 'ABOUT'.
    \end{error_analysis_gtbox}
    \begin{error_analysis_captionbox}{Caption by Qwen2.5-Omni-7B}
    The video opens with a young girl, Mackenzie, sitting on a stage. She is wearing a colorful outfit and a party hat. The text "MACKENZIE'S SOLO DANCE STYLE: ACRO" appears on the screen. Mackenzie then begins her dance routine, performing various acrobatic moves. She does a handstand, a cartwheel, and a backbend. The audience, including a woman in the audience, claps and cheers for her. The backdrop behind her reads "NATIONAL TALENT COMPETITION." The music plays in the background, adding to the excitement of the performance. The video ends with a shot of the audience clapping and a woman in a black dress standing on a stage with the words "WATCH MORE VIDEOS" and "THE SHOW MUST GO ON... DANCE MOMS" displayed on the screen.
    \end{error_analysis_captionbox}
    \begin{error_analysis_analysisbox}{Error Analysis}{analysisbg}
    The baseline model exhibits incomplete transcription, completely omitting the spoken dialogue and on-screen interview speech (e.g., the mother's remarks and audience exclamations), thereby failing to capture crucial semantic knowledge from the audio modality.
    \end{error_analysis_analysisbox}
    \begin{error_analysis_trainedbox}{Caption by AVSCap-7B(ours)}
    The scene opens on a dimly lit stage where a young girl stands in profile, wearing a vibrant two-piece dance outfit with pink, green, and yellow ruffles, topped with a yellow party hat featuring a pink pom-pom. On-screen text reads "MACKENZIE'S SOLO DANCE STYLE: ACRO". An upbeat pop song with a driving beat and female vocals singing, "Watch me do my dance, set the rules down, need to call the boys, welcome around," begins to play (Music). Mackenzie raises her arms and executes a handstand, followed by a front walkover. The camera briefly cuts to a woman in the audience with dark hair, wearing a black top and holding a blue pen, watching intently. Returning to the stage, the backdrop is now visible, displaying "NATIONAL TALENT COMPETITION" in large letters with star graphics. Mackenzie performs a backbend, lowering herself to the floor, then rolls onto her stomach, sits up, and completes a backward roll before running forward with a smile. The music continues with lyrics, "wanna have a good time, is it such a crime, over all the drama, let's go have some fun" (Music). The view shifts to the audience, showing a woman in a red top smiling broadly next to a young girl. The scene then transitions to an interview setting with the woman, now wearing a beige jacket and a statement necklace, positioned in front of a rack of colorful costumes. On-screen text identifies her as "MELISSA DANCE MOM". She explains, "Mackenzie is seven, and before she just did it for fun, and now I think she wants to really be good." (Speech). The video cuts back to the stage as Mackenzie performs a leaping jump, landing gracefully. The camera briefly shows Melissa in the audience again, laughing joyfully. On stage, Mackenzie continues her routine, executing a one-handed cartwheel as the music's chorus repeats, "party starts right now" (Music). The camera cuts to two women in the audience, one in a white jacket and the other in a sparkly black top, both laughing and enjoying the performance. Back on stage, Mackenzie performs a chin stand, her legs arched over her head, before rolling onto her stomach and sliding into a final seated pose with one hand resting on the floor. As she strikes her final pose, loud applause and cheering erupt from the audience (SFX). The camera shows Melissa clapping enthusiastically and exclaiming, "Oh, she did so good!" (Speech), while the woman in the sparkly black top also claps with a wide smile. The video concludes with a static end screen featuring a blue curtain backdrop. A woman dressed in black stands in the center, posing with one hand on her hip and one foot resting on scattered red rose petals. Text on the screen reads "WATCH MORE VIDEOS" on the left, and "THE SHOW MUST GO ON... DANCE MOMS" on the right, with "SUBSCRIBE" and "ABOUT" positioned below.
    \end{error_analysis_trainedbox}
    
\end{erroranalysiscontainer}

\begin{erroranalysiscontainer}{Example 3}

    \setlength{\tabcolsep}{0pt} 
    \renewcommand{\arraystretch}{0}
    \begin{tabularx}{\linewidth}{@{}XXXX@{}}
        \includegraphics[width=\linewidth]{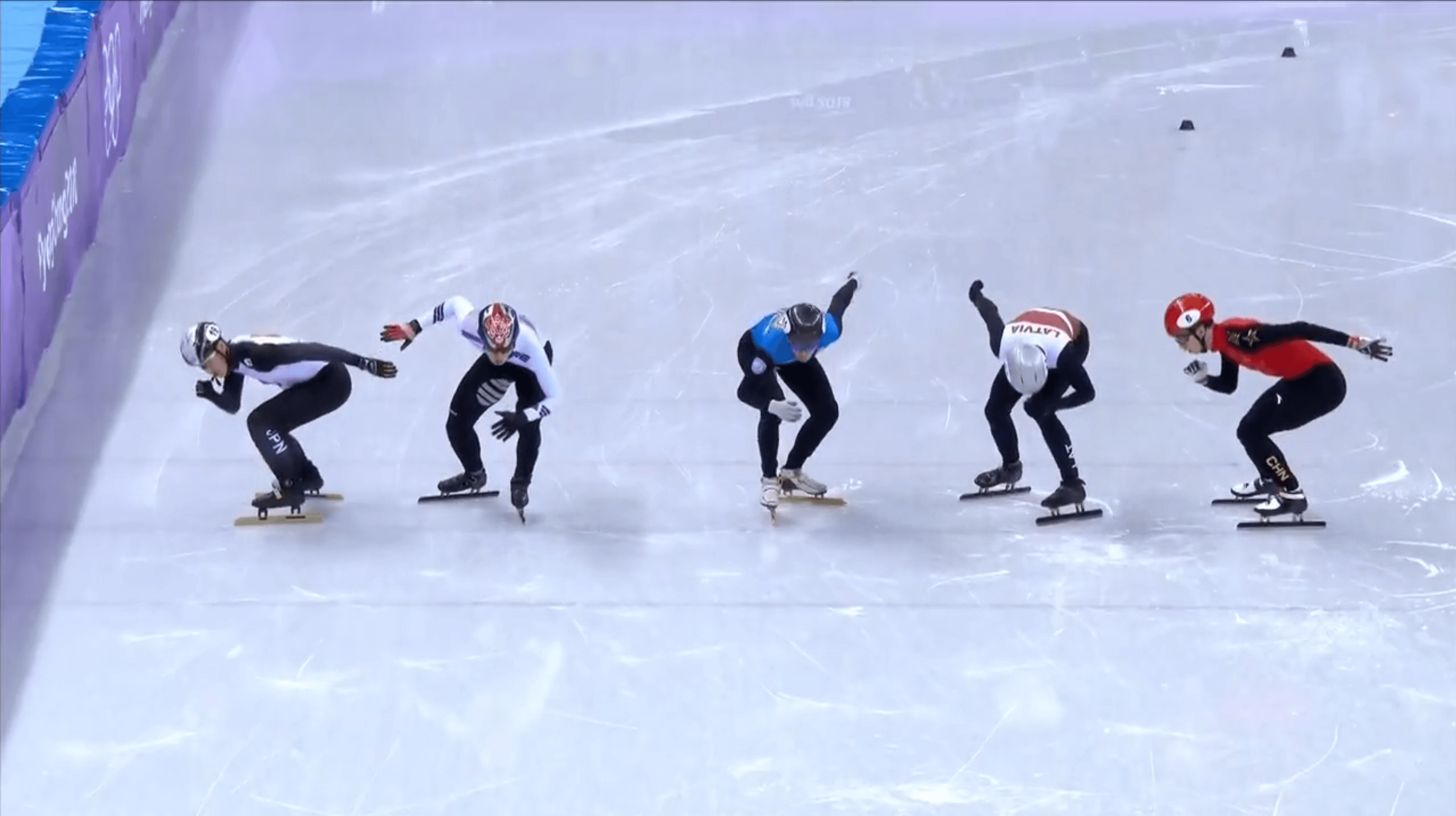} & \includegraphics[width=\linewidth]{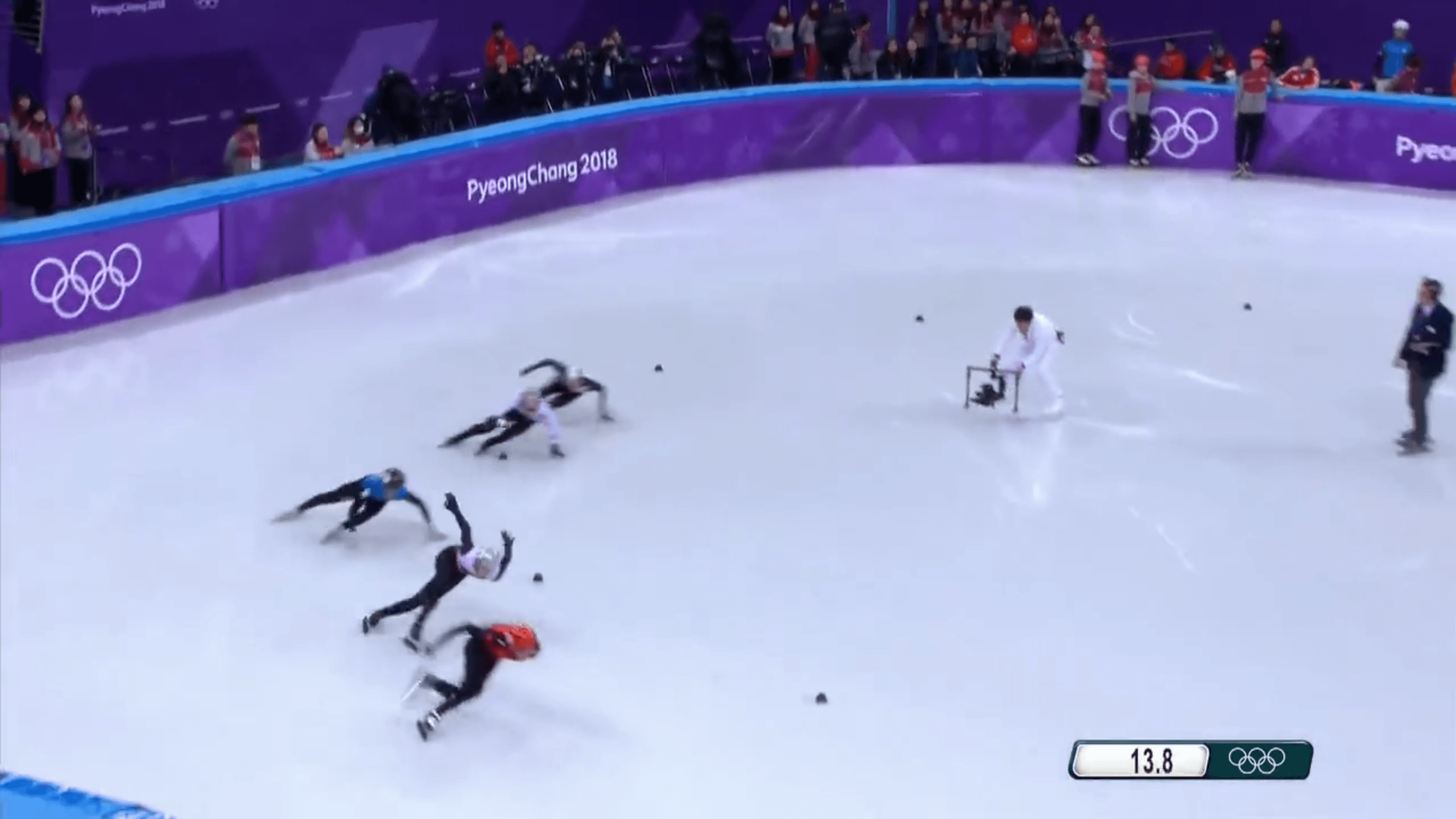} & \includegraphics[width=\linewidth]{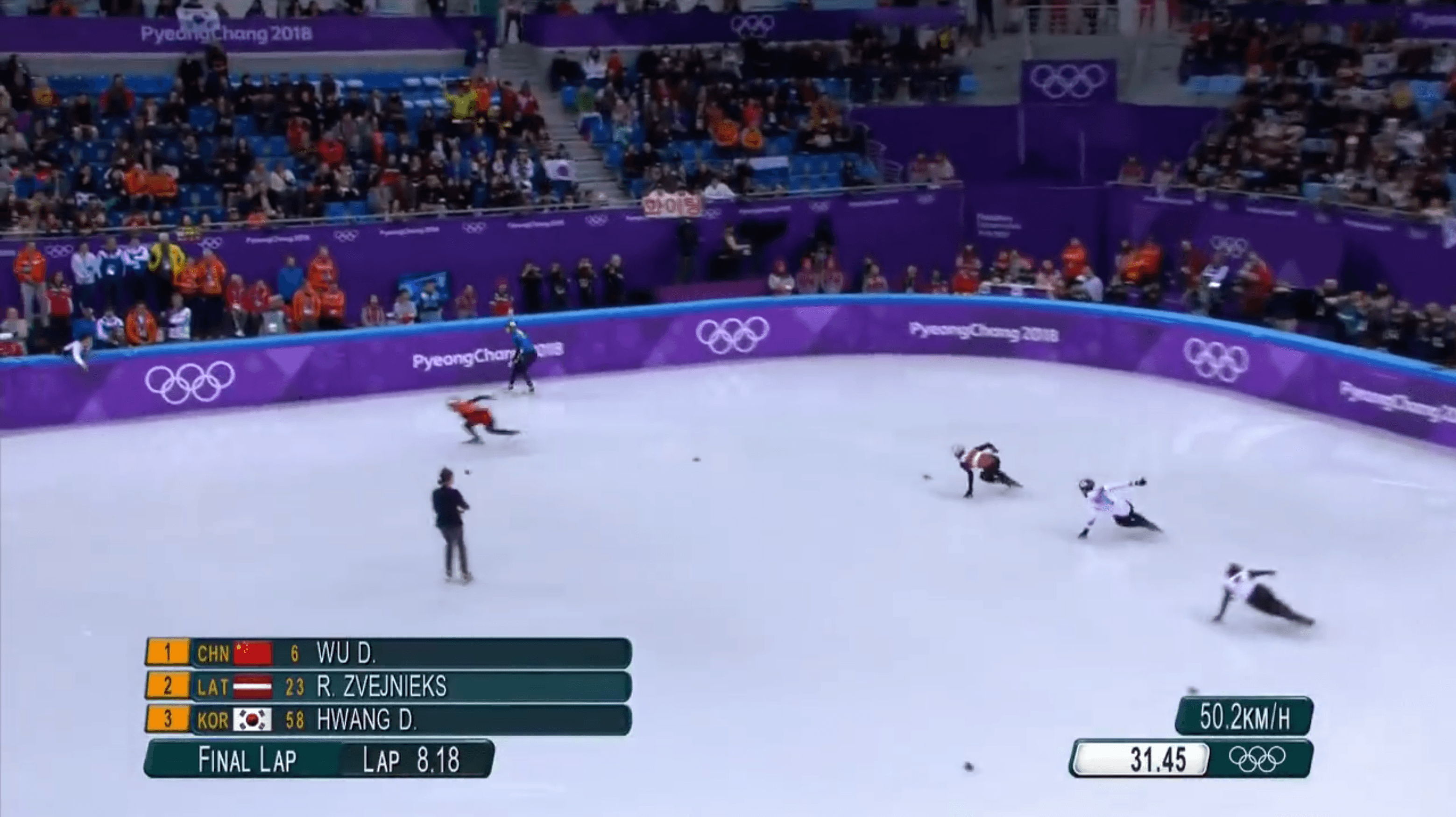} & \includegraphics[width=\linewidth]{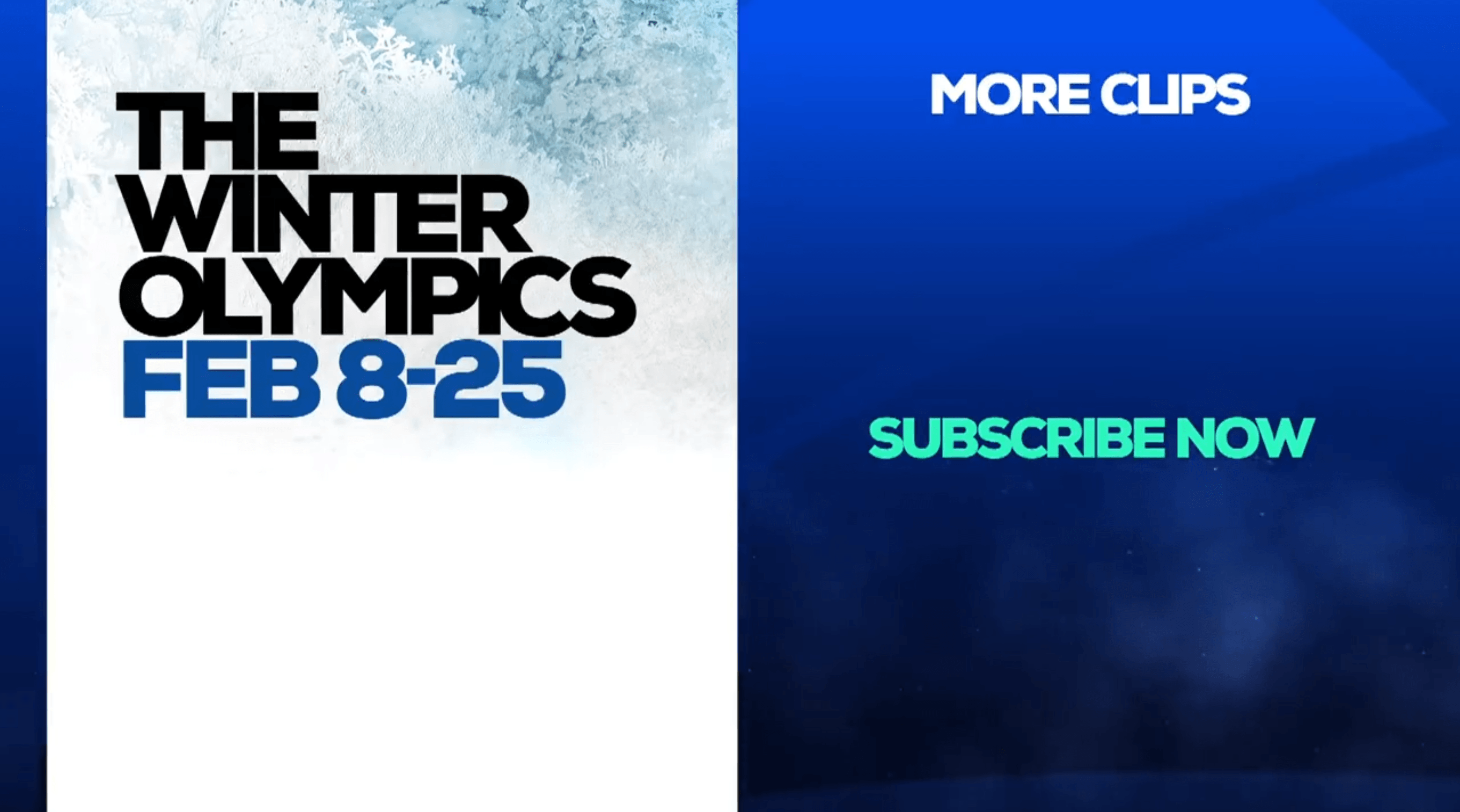} \\
    \end{tabularx}
    \vspace{2pt}

    \small \textbf{Prompt:} \\
    \textit{Please describe all the information in the video without sparing every detail in it. As you describe, you should also describe as much of the information in the audio as possible, and pay attention to the synchronization between the audio and video descriptions.}

    \vspace{2pt}
    \begin{error_analysis_gtbox}{Ground Truth}
    A high-angle view shows a short track speed skating starting line on an indoor ice rink where five skaters are lined up horizontally, wearing distinct aerodynamic suits and helmets. From left to right, they are dressed in black with white stripes, white with red and blue accents, blue with black, white with red, and red with black. Initially, the athletes stand upright and relaxed, looking down at the ice markings as the official says authoritatively, "Ready." (Speech). Upon hearing the command, they remain still for a split second. Suddenly, a starting pistol fires with a loud gunshot (SFX), and all five skaters simultaneously drop their bodies low, bending their knees deeply and leaning their torsos forward into a crouched, ready position, poised to explode off the starting line. The five skaters launch from their crouched starting positions as the commentator notes in a professional and observant tone, "And they're away cleanly for" (Speech). They drive their legs outward and swing their arms vigorously to gain momentum with the rhythmic scraping and whooshing of skates on ice (SFX). As the pack accelerates, a digital timer graphic in the bottom right corner counts up, displaying '2.2'. The camera pans smoothly to the left, tracking a tight pack of four skaters as they round a curve while the commentator states in an informative and excited tone, "500 meters four and a half laps around the track." (Speech). Leading the pack is a skater wearing a red helmet and a black suit with red shoulders representing China, leaning deeply with his left hand grazing the ice surface for balance. As they straighten out onto the track, a digital graphic overlay displays "4 LAPS TO GO" as the commentator excitedly says, "And they will be flying. Four laps to go now look at that speed 43.3 kilometers." (Speech). The skaters execute powerful crossover strides, their bodies low and aerodynamic, creating rhythmic, sharp scraping sounds of blades cutting into the ice (SFX). The commentator descriptively notes, "Wu Dajing of China has the lead right behind him..." (Speech) against the backdrop of purple rink barriers and a blurred crowd of spectators, accompanied by the continuous ambient roar from the stadium crowd (SFX). A leaderboard graphic displays the current race order: '1 CHN WU D', '2 LAT R. ZVEJNIEKS', and '3 JPN K. WATANABE', alongside a status indicator reading '3 Laps to Go' and a speed of '50.9 KM/H'. As the skaters round a curve, a racer in the back of the pack loses his footing and crashes, sliding helplessly across the rink into the purple padded barriers as the commentator states professionally, "And we have one racer down and out, the Kazakhstan is out." (Speech). The remaining three skaters continue their momentum down the straightaway to the continuous crowd cheering (SFX). A graphic overlay displays "2 Laps to Go." The commentator excitedly announces, "But Wu Dajing has opened up a sizable lead over Roberts Zvejnieks." (Speech). The commentator says in an anticipatory tone, "Korean racer Hwang Dae-heon trying to catch him." (Speech). Approaching the finish line, the leader extends his right skate forward in a lunge amidst crowd cheering (SFX), while the trailing skater simultaneously lunges to close the gap. They cross the line nearly together as the commentator declaratively states, "Photo finish." (Speech). A digital timer reads 39.81 as the race concludes. A results graphic lists Wu Dajing first, Hwang Daeheon second, and R. Zvejnieks third. The commentator emphatically declares, "But Wu Dajing, the definite winner of that heat" (Speech). The video concludes with a static promotional graphic for the Winter Olympics, displaying 'THE WINTER OLYMPICS' with the dates 'FEB 8-25' and a 'SUBSCRIBE NOW' call-to-action.
    \end{error_analysis_gtbox}
    \begin{error_analysis_captionbox}{Caption by Qwen2.5-Omni-7B}
    The video opens with a group of speed skaters on an ice rink, all wearing helmets and racing suits. They are lined up, ready to start a race. The skaters begin to move, gliding swiftly across the ice. The camera captures their speed and agility as they navigate the track. The crowd in the stands is visible, cheering and watching intently. The on-screen text shows the names and rankings of the skaters, with "WU D." in the lead. The speed of the skaters is displayed, reaching up to 50.9 km/h. As the race progresses, the skaters continue to push themselves, with some making sharp turns and others maintaining a steady pace. The final lap approaches, and the skaters are neck and neck, with "WU D." still in the lead. The video ends with a close-up of the skaters crossing the finish line, and the on-screen text confirms "WU D." as the winner. The video then transitions to a screen promoting "The Winter Olympics" with the dates "Feb 8-25" and a call to action to "Subscribe Now."
    \end{error_analysis_captionbox}
    \begin{error_analysis_analysisbox}{Error Analysis}{analysisbg}
    The baseline model fails to capture temporal alignment, listing isolated visual actions (such as the skaters starting and crossing the finish line) without syncing them with the auditory triggers, such as the starting pistol gunshot (SFX) or the commentator's live voiceover (Speech).
    \end{error_analysis_analysisbox}
    \begin{error_analysis_trainedbox}{Caption by AVSCap-7B(ours)}
    The video begins on an indoor ice rink with five speed skaters lined up at the starting line in a crouched position. From left to right, the skaters wear black, white, blue, white, and red suits. A voice commands, "Ready" (Speech), followed immediately by the sharp crack of a starting pistol (SFX). The skaters launch forward, their blades scraping the ice (SFX). An announcer states, "And they're away cleanly for this heat number two of the men's 500 meters..." (Speech) while a timer graphic in the bottom right corner begins counting up. As the skaters navigate the first curve, the skater in red takes the lead, with the announcer continuing, "And they will be flying. Four laps to go now. Look at that speed, 43.3 kilometers" (Speech) as a ranking overlay appears on the bottom left. The skaters execute powerful crossover strides, creating a rhythmic scraping sound of blades cutting into the ice (SFX). The commentator excitedly notes, "Wu Dajing of China has the lead right behind him..." (Speech) amidst the continuous ambient roar of the cheering crowd (SFX). As the skaters round a curve, a racer in the back of the pack loses his footing and crashes, sliding helplessly into the padded barriers (SFX). The commentator notes, "And we have one racer down and out, the Kazakhstan is out." (Speech). The remaining three skaters maintain high speed down the straightaway to the crowd's loud cheering (SFX). Approaching the final lap, the leader maintains a narrow lead over his opponent. He extends his right skate forward in a lunge to cross the finish line (SFX) as the commentator declares, "Photo finish." (Speech). A final results graphic lists Wu Dajing first, while the commentator states, "But Wu Dajing, the definite winner of that heat" (Speech). The video concludes with a static Winter Olympics promotional screen (Music).
    \end{error_analysis_trainedbox}

\end{erroranalysiscontainer}

\section{Details of Human Annotation for AVSCapBench}
\label{sec:annotation_details}

In this section, we provide detailed information regarding the human annotation process for AVSCapBench, directly addressing the ARR Responsible NLP Research requirements (D1--D4).

\subsection{Annotator Instructions and Workflow (D1)}
To ensure high-quality annotations, annotators were provided with the required pre-segmented video clips alongside a comprehensive annotation guideline. The workflow was straightforward: annotators watched the provided clips and wrote omni-modal captions strictly following our detailed standards. They were not responsible for any upstream video segmentation or downstream caption merging. 

The core instructions required annotators to accurately describe visual actions, comprehensively record auditory events (including speech, sound effects, and music), and explicitly use temporal conjunctions (e.g., ``while'', ``accompanied by'') to bind corresponding cross-modal events. After drafting a caption, annotators were required to cross-reference a predefined verification checklist to ensure the completeness of these three dimensions. Before entering the formal annotation phase, all participants underwent a qualification test where they annotated several sample videos and received feedback to ensure they fully grasped the required standards. Since the task only involves describing everyday videos, there were no risks of exposure to harmful materials, and a standard disclaimer was provided prior to the task.

\subsection{Recruitment and Payment (D2)}
We recruited 5 human annotators for this project. All annotators are graduate-level students with high English proficiency, ensuring the high linguistic quality of the captions. 

Annotators were compensated based on the time spent on the task. On average, annotating a single video clip took approximately 20 minutes. The payment was set at approximately \$12 USD per hour (which equates to roughly \$4 USD per clip). This compensation rate is well above the local statutory minimum wage and is considered highly competitive and adequate for data annotation tasks in the annotators' demographic region.

\subsection{Data Consent (D3)}
There are two aspects of data consent in our study. First, regarding the source videos: all video clips used to construct AVSCapBench were sourced from publicly available platforms (e.g., YouTube, TikTok) and existing open-source datasets. They are used strictly under fair use for academic research. We do not distribute the original video files; instead, we release the public URLs and timestamps. Second, regarding the annotators: before starting the task, all annotators were informed about the purpose of the research. They explicitly consented that their generated text annotations would be open-sourced and freely available to the academic community.

\subsection{Ethics Review Board Approval (D4)}
The data collection and annotation protocol was reviewed in accordance with our institution's ethical guidelines. Because the annotation task strictly involves describing publicly available, non-harmful video content and does not involve collecting any personally identifiable information (PII), psychological profiling, or exposure to offensive materials, the protocol was determined to be exempt from formal Ethics Review Board (IRB) approval.

\clearpage

\end{document}